\documentclass[runningheads]{llncs}

\usepackage{eccv}

\usepackage{tikz}
\usepackage{comment}
\usepackage{amsmath,amssymb} 
\usepackage{color}

\usepackage{eccvabbrv}

\usepackage{graphicx}
\usepackage{booktabs}

\usepackage[accsupp]{axessibility}  

\usepackage[pagebackref,breaklinks,colorlinks]{hyperref}

\usepackage{orcidlink}

\usepackage{amsmath,amsfonts,bm}

\def\eqref#1{equation~\ref{#1}}

\def\floor#1{\lfloor #1 \rfloor}
\def\1{\bm{1}}

\newcommand{\test}{\mathcal{D_{\mathrm{test}}}}

\DeclareMathAlphabet{\mathsfit}{\encodingdefault}{\sfdefault}{m}{sl}
\SetMathAlphabet{\mathsfit}{bold}{\encodingdefault}{\sfdefault}{bx}{n}

\newcommand{\Co}{\mathrm{Co}}

\usepackage{hyperref}
\usepackage{url}
\usepackage[T1]{fontenc}

\usepackage{mathabx}
\usepackage{amsmath}
\usepackage{subcaption}
\usepackage{graphicx}
\usepackage{wrapfig, booktabs}
\usepackage{algorithm}
\usepackage{algpseudocode}
\newcommand \ignore[1]{}
\algrenewcommand\algorithmicrequire{\textbf{Input:}}
\algrenewcommand\algorithmicensure{\textbf{Output:}}

\begin{document}

\title{Improving Robustness to Model Inversion Attacks via Sparse Coding Architectures}
\title{Improving Robustness to Model Inversion Attacks via Sparse Coding Architectures}
\title{Improving Robustness to Model Inversion Attacks via Sparse Coding Architectures}
\titlerunning{Improving Robustness to MI Attacks via Sparse Coding Architectures}

\author{Sayanton V. Dibbo\inst{1, 2}\orcidlink{0000-0002-8461-6966} 
\and
Adam Breuer\inst{1}\orcidlink{0000-0002-5978-1070} 
\and
Juston Moore\inst{2}\orcidlink{0000-0003-2515-3647}
\and
Michael Teti\inst{2}\orcidlink{0000-0002-0754-1761}}
\authorrunning{S. Dibbo et al.}
\institute{Dartmouth College, Hanover, NH 03755, USA \and
Los Alamos National Laboratory, Los Alamos, NM 87545, USA
\email{\{f0048vh,adam.breuer\}@dartmouth.edu}\\
\email{\{jmoore01,mteti\}@lanl.gov}}
\maketitle







\begin{abstract}
Recent model inversion attack algorithms permit adversaries to reconstruct a neural network's private and potentially sensitive training data by repeatedly querying the network. In this work, we develop a novel network architecture that leverages sparse-coding layers to obtain superior robustness to this class of attacks. Three decades of computer science research has studied sparse coding in the context of image denoising, object recognition, and adversarial misclassification settings, but to the best of our knowledge, its connection to state-of-the-art privacy vulnerabilities remains \mbox{unstudied}. In this work, we hypothesize that sparse coding architectures suggest an advantageous means to defend against model inversion attacks because they allow us to control the amount of \mbox{irrelevant} private information encoded by a network in a manner that is known to have little effect on classification accuracy. 
Specifically, compared to networks trained with a variety of state-of-the-art defenses, our sparse-coding architectures maintain comparable or higher classification accuracy while degrading state-of-the-art training data reconstructions by factors of $1.1$ to $18.3$ across a variety of reconstruction quality metrics (PSNR, SSIM, FID).
This performance advantage holds across $5$ datasets ranging from CelebA faces to medical images and CIFAR-10, and across various state-of-the-art SGD-based and GAN-based inversion attacks, including \emph{Plug-\&-Play} attacks. We provide a cluster-ready PyTorch codebase to promote research and standardize defense evaluations.
\end{abstract}



%
%

\section{Introduction}
\label{sec:intro}
The popularization of machine learning has been accompanied by the widespread use of neural networks that were trained on private, sensitive, and proprietary datasets. This has given rise to a new generation of privacy attacks that seek to infer private information about the training dataset simply by inspecting the representation of the training data that remains encoded in the model’s parameters~\cite{fredrikson2015model,gong2016you,kariyappa2021maze,zhong2022understanding,mehnaz2022your,wang2022dualcf,yuan2022attack,hu2022membership,zhang2023apmsa,sanyal2022towards,struppek2022plug,carlini2023extracting,dibbo2023model,li2023sok, dibbo2023sok}.

Of particular concern is a devastating stream of privacy attacks known as model inversion. Model inversion attacks leverage the network’s parameters or classifications in order to reconstruct entire images or data that were used to train the network. Early work on model inversion focused on a white-box setting where the attacker has unfettered access to the model or auxiliary information about the training data~\cite{fredrikson2015model,hitaj2017deep,wang2019beyond,zhang2020secret,wei2020framework}. However, recent work has shown that standard network architectures are vulnerable to model inversion attacks even when attackers have no knowledge of the model’s architecture or parameters, and only have access to the model's classifications or its intermediate outputs, such as leaked outputs from a single hidden network layer~\cite{yang2019neural,mehnaz2022your,salem2020updates,melis2019exploiting,an2022mirror,gong2023gan}.

\centerline{\emph{Are different network architectures robust to model inversion attacks?}}

Such attacks are feasible because each hidden layer of a standard network architecture captures a detailed representation of the training data. It is well-known that standard dense layers exhibit a tendency to memorize their inputs~\cite{haim2022reconstructing,carlini2022privacy,rigaki2020survey}, so even a minimal leak of a network's class distribution output or a leak of its intermediate outputs from a single layer is often sufficient to train an inverse mapping for data reconstruction. More concretely, state-of-the-art inversion attacks work by submitting externally obtained images to the model, observing leaked outputs, then using this data to train a new `inverted’ neural network that reconstructs (predicts) an input image given a leaked output. This can be accomplished either directly via SGD, or by optimizing a GAN, and we consider both approaches here. Such attacks on standard network architectures can reconstruct private training images that are clearly recognizable by humans familiar with the training data~\cite{hitaj2017deep,yang2019neural,he2019model,wei2020framework,aivodji2019gamin,kahla2022label,struppek2022plug,gong2023gan, fang2023gifd}.

\sloppy

Recent work has pursued a diverse array of defense strategies to mitigate these attacks. For example, \cite{gong2023gan} augments the training dataset with GAN-generated fake samples designed to inject spurious features into the trained network that mislead the gradients computed during inversion attacks. In contrast, multiple recent defenses add regularization terms during training that attempt to penalize training data memorization \cite{peng2022bilateral, wang2021improving}. Other recent defenses noise the network weights to obfuscate memorized data \cite{titcombe2021practical, abuadbba2020can, mireshghallah2020shredder}, or noise and clip training gradients via DP-SGD \cite{hayes2023bounding}. All such approaches are costly: data augmentation-based defenses entail the computational burden of building a GAN and applying sophisticated parameter tuning techniques during training; regularization-based defenses explicitly trade away classification accuracy for less memorization, and noise-based defenses are also known to impose significant accuracy costs. Until very recently, there were no known provable guarantees for model inversion defenses, and the current best-known guarantees require a DP-SGD-based training algorithm that imposes a significant computational burden and accuracy loss to obtain privacy guarantees that are impractically weak for these attacks \cite{hayes2023bounding}.

Very little is known about how a network’s architecture contributes to its robustness (or vulnerability). This is surprising since throughout three decades of research in other domains such as image denoising~\cite{barlow1961coding,field1994goal,chen2001atomic,olshausen2004sparse,candes2004new,rozell2008sparse,krause2010submodular,ahmad2019can}, object recognition \cite{olshausen1995sparse,schneiderman2004feature,kavukcuoglu2010fast,hannan2023mobileptx}, and adversarial misclassification \cite{sun2019adversarial,paiton2020selectivity,kim2020modeling,teti2022lcanetst, dibbo2023lcanets++}, researchers seeking to control their model’s representations of the data have heavily studied sparse coding-based architectures that prune unnecessary details and preserve only the information that is essential to the model objective. Specifically, sparse coding seeks to approximately represent an image (or layer) with only a small set of basis vectors selected from an overcomplete dictionary \cite{field1994goal,olshausen2004sparse,candes2004new}. While it is well-known that computing a sparse representation using a standard objective function is NP-hard in general \cite{natarajan1995sparse,davis1997adaptive,jiang2012submodular}, we now benefit from fast approximation algorithms that efficiently compute high-quality sparse representations \cite{lee2006efficient,rozell2008sparse,kavukcuoglu2010fast,krause2010submodular,jiang2012submodular,mirzasoleiman2015lazier,breuer2020fast,chen2021best}. Sparse coding architectures leverage this technique by inserting a sparse network layer after a dense layer, such that the sparse layer reduces the dense layer’s outputs to a sparse representation. 

To our knowledge, sparse coding architectures have not been studied in the context of model inversion or privacy attacks. However, they suggest an advantageous means to prevent such attacks because they control the amount of irrelevant private information encoded in a model’s intermediate representations in a manner that is known to have little effect on its accuracy, that can be computed efficiently during training, and that adds little to the trained model’s overall parameter complexity. Put simply, sparsifying a network's representations is a natural means to preclude memorization of detailed information about its inputs that is unnecessary to obtain high accuracy, so even an idealized `perfect attacker' could only hope to recover a sparsified, un-detailed training image.


\textbf{Main contribution.} We begin by showing that an off-the-shelf sparse coding preprocessing step offers performance advantages compared to state-of-the-art data augmentation, regularization, and noise based defenses in terms of robustness to model inversion attacks. We then refine this idea into a network architecture that achieves superior performance. Our main result is a novel sparse-coding architecture, \textsc{SCA}, that is robust to state-of-the-art model inversion attacks.

\textsc{SCA} is defined by pairs of alternating sparse coded and dense layers that jettison unnecessary private information in the input image and ensure that downstream layers do not e.g., reconstruct this information. 
We show that \textsc{SCA} maintains comparable or higher classification accuracy while degrading state-of-the-art training data reconstructions $1.1$ to $11.7$ times more than $8$ state-of-the-art data augmentation, regularization, and noise-based defenses in terms of PSNR and FID metrics and $1.1$ to $720$ times more in terms of SSIM. This performance advantage holds across $5$ datasets ranging from CelebA faces to medical images and CIFAR-10, and across various state-of-the-art SGD-based and GAN-based inversion attacks, including \emph{Plug-\&-Play} attacks. \textsc{SCA}'s defense performance is also more stable than baselines across multiple runs. We emphasize that, unlike recent state-of-the-art defenses that require sophisticated parameter tuning to perform well, \textsc{SCA} obtains these results absent parameter tuning (i.e., using default sparsity parameters) because sparse coding naturally precludes networks from memorizing detailed representations of the training data. 

More broadly, our results show a deep connection between state-of-the-art ML privacy vulnerabilities and three decades of computer science research on sparse coding for other application domains. We provide a comprehensive cluster-ready PyTorch codebase to promote research and standardize defense evaluation.

\section{Threat models}
\label{sec:advsetting}
We consider three threat models that span the diverse range of powerful and well-informed attackers considered in recent work. We emphasize that a defense that performs well in all three settings provides strong evidence of its privacy protections under weaker, more realistic threat models with real-world attackers. 

\textbf{1. Plug-\&-Play threat model \cite{struppek2022plug}.} Plug-\&-Play attacks are considered the most performant recent attacks. These attacks optimize the intermediate representation of StyleGAN's input vectors so that generated images maximize the target network's class prediction probability, which the attacker can query.

\smallskip 

Separately, recent theoretical work on model inversion emphasizes that a strong threat model should capture `worst-case' attackers with direct access to the information-rich, high-dimensional intermediate outputs of the target model that store private information about the training data, as well as ideal training data examples for training an inverted model \cite{hayes2023bounding, abadi2016deep}. We consider two variants:

\textbf{2. End-to-end threat model.}  We consider an attacker with access to all of the \emph{last} hidden layer's raw, high-dimensional outputs, as well as a large number of ideal training data examples drawn from the true training dataset~\cite{wang2022group,song2021systematic}.

\textbf{3. Split network threat model (Federated Learning).}
We also consider the split network threat model described by \cite{titcombe2021practical}. There has been much recent interest in Federated Learning architectures that split the network across multiple agents \cite{konevcny2016federated,mcmahan2017communication,bonawitz2019towards, fang2023gifd, he2019model}, particularly for privacy-fraught domains such as medicine where legal requirements limit data sharing \cite{vepakomma2018split,kaissis2020secure}. These architectures are known to be susceptible to model inversion attacks,\cite{titcombe2021practical, he2019model, fang2023gifd}, and defenses are urgently needed. This threat model also allows us to capture a different view of a `worst-case' threat model: Model inversion attacks are known to be more effective when the attacker has access to outputs from earlier layers that may exhibit a more direct representation of the input images \cite{he2019model}. To capture this `worst-case', we consider the setting where the attacker has access to raw intermediate outputs from the \emph{first} linear network layer. As before, we also assume the attacker has access to ideal training data examples drawn from the actual training datasets. Appendix B provides additional details about the model inversion threat models.

\section{\textsc{SCA} architecture}
\label{sec:method}
We now describe the \textsc{SCA} architecture, which is defined by alternating pairs of Sparse Coding Layers (SCL) and dense layers, followed by downstream linear and/or convolutional layers. 

\begin{wrapfigure}{R}{6cm}
\begin{subfigure}{.45\textwidth}
  \centering
  \includegraphics[width=5.8cm]
  {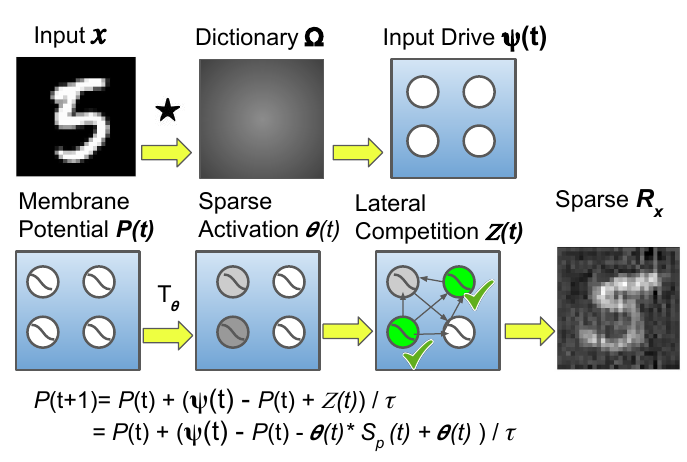}
\end{subfigure}
\caption{Pipeline of neuron (membrane potential) dynamics in Sparse Coding Layer (SCL) with lateral competitions.}
\label{fig:SCL_principle}
\end{wrapfigure}

\paragraph{\textbf{Sparse Coding Layer (SCL).}}
\label{subsection:scl_details}
Sparse coding converts raw inputs to sparse representations where only a few neurons whose features are useful in reconstructing the inputs are active. Our Sparse Coding Layer (SCL) performs sparse coding to obtain a sparse representation of a previous dense layer's representation (if the SCL is not the first layer in the network) or of the inputs (if the SCL is the first layer in the network). Fig. \ref{fig:SCL_principle} illustrates the working principle of SCL.


Formally, each SCL performs a reconstruction minimization problem to compute the sparse representation of its inputs (either a previous layer's representation or of the inputs to the network). Suppose the input to a (2D convolutional) SCL is $\mathcal X \in \mathbb{R} ^{\mathcal C \times \mathcal H \times \mathcal W}$ with $\mathcal H$ height, $\mathcal W$ width, and $\mathcal C$ channels/features. The goal is to find the sparse representation $\mathcal R_{x} \in \mathbb{R}^{\mathcal F \times \floor {\mathcal H /S_{h}} \times \floor{\mathcal W/S_{w}}}$, where $\mathcal R_x$ has few active neurons and corresponds to a denoised version of the input $\mathcal X$, and $S_{w}$ and $S_{h}$ indicate convolutional strides across the width and height of the input, respectively. $F$ is the number of convolutional features in the SCL layer's dictionary, $\Omega \in \mathbb{R} ^{\mathcal F \times \mathcal C \times \mathcal H_f \times \mathcal W_f}$, where $\mathcal H_f$ and $\mathcal W_f$ are the height and width of each convolutional feature, respectively. Per Figure~\ref{fig:SCL_principle}, the sparse coding layer starts with its input, $\mathcal X$, and dictionary of features, $\Omega$, to produce $\mathcal R_x$ by solving the following sparse reconstruction problem:
\begin{equation}\label{eqn:re_loss}
\min_{\mathcal R_x} \frac{1}{2} || \mathcal X- \mathcal R_{x} \circledast \Omega ||_2^2 + \lambda || \mathcal R_{x} ||_1
\end{equation}
\noindent where the first term represents how much information is preserved about $\mathcal X$ by $\mathcal R_x$ by measuring the difference between $\mathcal X$ and its reconstruction, $\mathcal R_x \circledast \Omega$, computed with a transpose convolution, $\circledast$. The second term measures how sparse $\mathcal R_x$ is, and $\lambda$ is a constant which determines the trade-off between reconstruction fidelity and sparsity. Equation~\ref{eqn:re_loss} is convex in $\mathcal R_{x}$, meaning we will always find the optimal $\mathcal R_x$ that solves Equation \ref{eqn:re_loss}.

Among different techniques to perform sparse coding, we leverage the commonly used Locally Competitive Algorithm (LCA) \cite{rozell2008sparse}. LCA implements a recurrent network of leaky integrate-and-fire neurons that incorporates the general principles of thresholding and feature-similarity-based competition between neurons to solve Equation \ref{eqn:re_loss}. While Rozell introduced LCA in the non-convolutional setting, it can be readily adapted to the convolutional setting (see Appendix A.2 for details). Specifically, each LCA neuron has an internal membrane potential $\mathcal P$ which evolves per the following differential equation:
\begin{equation}\label{eqn:ode_potential}
\mathcal {\dot P}(t) = \frac{1}{\tau}[\Psi (t) - \mathcal {P}(t) - \mathcal R_x(t) * \mathcal G]
\end{equation}
where $\tau$ is a time constant, $\Psi (t) = \mathcal X * \Omega$ is the neuron's bottom-up drive from the input computed by taking the convolution, $*$, between the input, $\mathcal X$, and the dictionary, $\Omega$, and $- \mathcal P (t)$ is the leak term \cite{teti2022lcanetst,kim2020modeling}. Lateral competition between neurons is performed via the term $- \mathcal R_x (t) * \mathcal G$, where $\mathcal G = \Omega * \Omega - I$ is the similarity between each feature and the other $\mathcal F$ features ($-I$ prevents self interactions). $\mathcal R_x$ is computed by applying soft threshold activation $T_{\lambda}(x) = \textnormal{relu}(x - \lambda)$ to the neuron's membrane potential, which produces nonnegative, sparse representations. Overall, this means that in LCA neurons will compete to determine which ones best represent the input, and thus will have non-zero activations in $\mathcal R_x$, the output of the SCL that is passed to the next layer. 

\begin{wrapfigure}{RB}{7.41cm}
\begin{subfigure}{.4\textwidth}
  \centering
  \includegraphics[width=7.41cm, height=2.5cm]
  {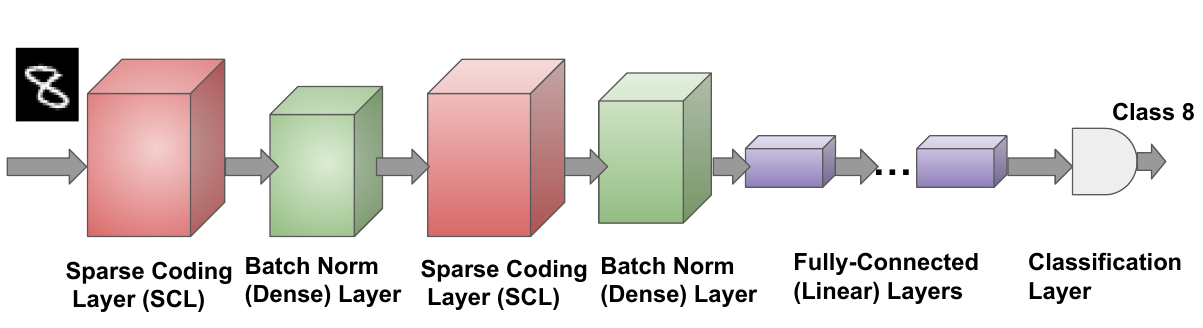}
\end{subfigure}
\caption{Architecture of \textsc{SCA}.}
\label{fig:SG_principle}
\end{wrapfigure}

\paragraph{\textbf{\textsc{SCA} architecture}.}
\label{subsection:sguard_details}
The \textsc{SCA} architecture is defined by the use of multiple \emph{pairs of sparse coding and dense (batch norm) layers} after the input image, which can then be followed by other (linear, convolutional) layers. Fig.~\ref{fig:SG_principle} illustrates this design principle. The \emph{key intuition} is that the first sparse layer jettisons unnecessary private information in the input image. Then, by alternating sparse-dense pairs of layers, we ensure that unnecessary information is also jettisoned from downstream layers. In this manner, downstream layers also do not convey unnecessary private information to the adversary, and they also do not e.g.  learn to reconstruct private information jettisoned by the first sparse layer. In short, previous defenses work by trying to mislead attackers by pushing model features in a wrong direction, either randomly via noise or strategically via adversarial examples, or by disincentivizing memorization during training. In contrast, \textsc{SCA} directly removes the unnecessary private information. Training \textsc{SCA} is identical to training a standard network with one exception: after each backprop updates non-sparse layers, we perform a fast update on the sparse layers, except for the first sparse layer that sparse-codes the image input.
\footnote{We can optionally also perform a backpropagation on sparse layers after updating them each iteration via sparse coding. We do this in our experiments.}



\paragraph{\textbf{\textsc{SCA} training complexity \& large scale applications.}}
\label{subsec:spg_complexity}
While we focus on the neuron lateral competition approach to sparse coding as it is practically convenient and well-represented in recent work \cite{teti2022lcanetst}, we note that for large-scale machine learning applications, we now have practical parallel algorithms that learn the sparse coding dictionary near-optimally w.p. in parallel time (adaptivity) that is logarithmic in the size of the data \cite{jiang2012submodular,breuer2020fast,chen2021best}. Fast single-iteration heuristics are also available (see e.g. \cite{wu2020fast}).   Thus, even for large-scale applications, computing sparse representations while training \textsc{SCA} adds little computational overhead compared to sophisticated optimization-based techniques necessary for recent defenses \cite{gong2023gan}. In practice, even our basic sparse coding research implementations (see Section \ref{sec:exp} and Appendix A.3) are slightly faster than optimized Torch implementations of the best-performing recent defense \cite{peng2022bilateral}. \label{fast_sparse}

\section{Experiments}
\label{sec:exp}

xOur goal in this section is to show that \textsc{SCA} performs well compared to state-of-the-art defenses as well as practical defenses used in leading industry models in terms of both classification accuracy and various reconstruction quality metrics. To evaluate its performance comprehensively, we test \emph{all combinations} of $5$ diverse datasets, $3$ threat models, $9$ defense baselines, plus multiple runs-per-experiment and various sparsity parameters $\lambda$. See Appendix A.1-A.6.

\label{subsec:dataset_details}
\paragraph{\textbf{Five benchmark datasets.}} We test our performance on \emph{all} \textbf{5} diverse datasets used to benchmark model inversion attacks across the recent literature: \textbf{CelebA} hi-res RGB faces, \textbf{Medical MNIST} medical images, \textbf{CIFAR-10} hi-res RGB objects,  \textbf{MNIST} grayscale digits, and \textbf{Fashion MNIST} grayscale objects.


\paragraph{\textbf{Three attacks.}}\label{subsec:bbox_setup}
For each dataset, we conduct three sets of experiments corresponding to our three threat models. First, we test SCA and baselines' defenses against a state-of-the-art \emph{Plug-\&-Play attack \cite{struppek2022plug} that leverages StyleGAN3} to obtain high-quality reconstructions. Second, we compare \textsc{SCA} networks to a variety of baselines in terms of their robustness to a state-of-the-art \emph{end-to-end network attack that leverages leaked raw high-dimensional outputs from the networks' last hidden layer, as well as held-out training data drawn from the true training dataset.} This allows us to assess \textsc{SCA}'s defenses in a realistic setting with a well-informed adversary. Our third set of experiments tests performance in a \emph{split network setting of \cite{titcombe2021practical} where the attacker has access to leaked raw outputs from the \emph{first} linear network layer}. Robustness in this setting is desirable because model inversion attacks are known to be more effective on earlier hidden layers \cite{he2019model}, and also because an algorithm that is robust to such attacks would be an effective defense under novel security paradigms such as Federated Learning, which is vulnerable to inversion \cite{titcombe2021practical} (see Appendix A.5).

\paragraph{\textbf{Nine defense baselines.}}
\label{subsec:model_details}
We compare \textsc{SCA} to $9$ baselines plus extra variants, including SOTA defenses and practical defenses used in leading industry models:
\begin{itemize}
    \item \textbf{No-Defense}. The baseline target model with no added defenses.
    \item \textbf{Hayes et al.} \cite{hayes2023bounding}. We train a DP-SGD defense that noises and clips gradients during training. This is the only defense with provable guarantees.  
    \item \textbf{Gong et al.} \cite{gong2023gan}. We train the very recent defense from \cite{gong2023gan} that uses sophisticated tuning and two types of GAN-generated images. We also try a `\textbf{++}' version that adds extra Continual Learning accuracy optimizations.  
    \item \textbf{Peng et al.} \cite{peng2022bilateral}. We train the Bilateral Dependency defense that adds a loss function for redundant input memorization during training.
    \item \textbf{Wang et al.} \cite{wang2021improving}. We train a Mutual Information Regularization defense that penalizes dependence between inputs and outputs during training.
    \item \textbf{Titcombe et al.} \cite{titcombe2021practical}. We train a state-of-the-art Laplace $\mathcal{L}(\mu$$=$$0,b$$=$$0.5)$ noise defense as in  \cite{titcombe2021practical}. We also try more noise---see Appendix D. 
     \item \textbf{Sparse-Standard}. We train an off-the-shelf sparse coding architecture \cite{teti2022lcanetst} with $1$ sparse layer after the input image via lateral competition as in \textsc{SCA}.
    \item \textbf{GAN [common industry defense]}. We train a GAN for $25$ epochs to generate fake samples, then train the target model with both original and GAN-generated samples. This defense is frequently used in industry.
    \item \textbf{Gaussian-Noise [common industry defense]}. We draw noises from \mbox{$\mathcal{N}(\mu$=$0$, $\sigma$=$0.5)$} and inject them into intermediate dense layers post-training.
\end{itemize}

\paragraph{\textbf{\textsc{SCA} \emph{without} parameter tuning.}}
\label{subsec:sparseguard0}
In all experiments, we consider the simplest case of \textsc{SCA} architecture that contains \textsc{SCA}'s alternating sparse-and-dense layer pairs followed by only linear layers. We note that adding downstream convolutional layers or more sophisticated downstream architectures is certainly possible, though we avoid this here in order to compare the essence of the \textsc{SCA} approach to the baselines. Appendix A.4 describes \textsc{SCA} details. In the split network setting, we are careful to use slightly shallower \textsc{SCA}  architectures with fewer linear layers to match the split network experiments of \cite{titcombe2021practical}.

Recent state-of-the-art defenses such as GAN-based defenses require sophisticated automatic parameter tuning techniques such as focal tuning and continual learning to obtain high performance \cite{gong2023gan}. To test whether \textsc{SCA} can be effective \emph{absent} parameter tuning, we just run \textsc{SCA} with 
sparsity parameter $\lambda$ set to $\mathbf{0.1}$, $\mathbf{0.25}$, or $\mathbf{0.5}$---the default values from various sparse coding contexts.



\paragraph{\textbf{Performance metrics.}}
\label{subsec:metrics}
We evaluate attackers' reconstructions using multiple standard metrics. 
Let $X^*_{in}$ denote the reconstruction of training image $X_{in}$. Then:
\begin{itemize}
    \item \textbf{Peak signal-to-noise ratio (PSNR)} [\emph{lower=better}]. PSNR is the ratio of max squared pixel fluctuations from $X_{in}$  to $X*_{in}$ over mean squared error. 
    \item \textbf{Structural similarity (SSIM)}~\cite{wang2004image} [\emph{lower=better}]. SSIM measures the product of luminance distortion, contrast distortion, \& correlation loss:\\$SSIM ({X_{in}}, {X^*_{in}}) = l_{dis}({X_{in}}, {X^*_{in}}) c_{dis}({X_{in}}, {X^*_{in}}) c_{loss}({X_{in}}, {X^*_{in}})$. \ 
    
    \item \textbf{Fréchet inception distance (FID)}~\cite{heusel2017gans} [\emph{higher=better}]. FID measures reconstruction quality as a distributional difference between $X_{in}$ and $X^*_{in}$:\\ 
    $ FID^2 ({X_{in}}, {X^*_{in}}) =||\mu_{X_{in}}-\mu_{X^*_{in}}||^2 + Tr (\Co_{X_{in}}+\Co_{X^*_{in}}-2*\sqrt{\Co_{X_{in}} \cdot \Co_{X^*_{in}}})$ 
\end{itemize}

\paragraph{\textbf{Target model.}}
\label{subsec:target_model}
We focus on privacy attacks on linear networks because they capture the essence of the privacy attack vulnerability ~\cite{fredrikson2015model,hidano2017model}, and because there is broad consensus that a principled understanding of their emerging privacy (and security) vulnerabilities\footnote{We also note that results on linear models may generalize better than results on more application-specific models, and linear models trained on private data remain ubiquitous among top industry products.} is urgently needed\cite{sannai2018reconstruction,liu2019socinf,wu2022membership,heredia2023adversarial}.

\paragraph{\textbf{PyTorch codebase, replicability, and evaluation standardization.}}
\label{subsec:exp_setup} For all experiments, we consider the standard train test split of $70$\% and
$30$\%. After training each defense model, we run attacks to reconstruct the entire training set and compare reconstruction performance. We run all the experiments on a standard industry production cluster with $4$ nodes and DELL Tesla V100 GPUs with $40$ cores. 
\textcolor{black}{ \emph{We provide a cluster-ready PyTorch codebase on our project page at:}} \href{https://sayantondibbo.github.io/SCA}{https://sayantondibbo.github.io/SCA}.



\subsection{Experimental results overview}
\label{results}
\textbf{\emph{Defense.}} Across the $3$ attack settings and $5$ datasets, \textsc{SCA} maintains comparable or higher classification accuracy while degrading state-of-the-art training data reconstructions $1.1$ to $11.7$ times more than the $9$ baselines in terms of PSNR \& FID, and $1.1$ to $720$ times more in terms of SSIM. This performance gap exceeds the scale of improvements made by recent algorithms. \textsc{SCA}'s defense is also more stable than baselines across multiple runs. This is because unlike for baselines, even an idealized `perfect attacker' can only hope to recover a sparsified, un-detailed training image from SCA. We show results here for the $2$ most privacy-sensitive datasets of medical images and CelebA in $3$ threat models, and defer the $9$ other \{dataset, attack\} combinations to Appendix C.

\smallskip
\noindent\textbf{\emph{Accuracy.}} Typically, obtaining greater defense means trading away accuracy. However, in $6$ of the $15$ experiments (MNIST + FashionMNIST)$\times$(Plug \& Play + end-to-end + split networks), SCA \emph{outperforms no-defense \textbf{and} all baselines' accuracy}. SCA also outperforms all baselines' accuracy on CelebA in Plug \& Play, and a sparse approach is within 0.003 of the best accuracy on an 8th experiment. \emph{No other single SOTA baseline wins on accuracy this consistently}. We emphasize that unlike baselines that do accuracy hyperparameter tuning, we obtain this result \emph{absent} any such tuning. SCA drops accuracy on MedMNIST (which is the most imbalanced \& has fewest training examples). However, tuning of SCA (kernel size 5 $\rightarrow$ 7) improves SCA's accuracy on MedMNIST in Table 3 from 94.6\% to 97\%—See Appendix G and Table 12.

\begin{figure*}[b]
\centering
  \centering
  \includegraphics[width=11.2cm]
  {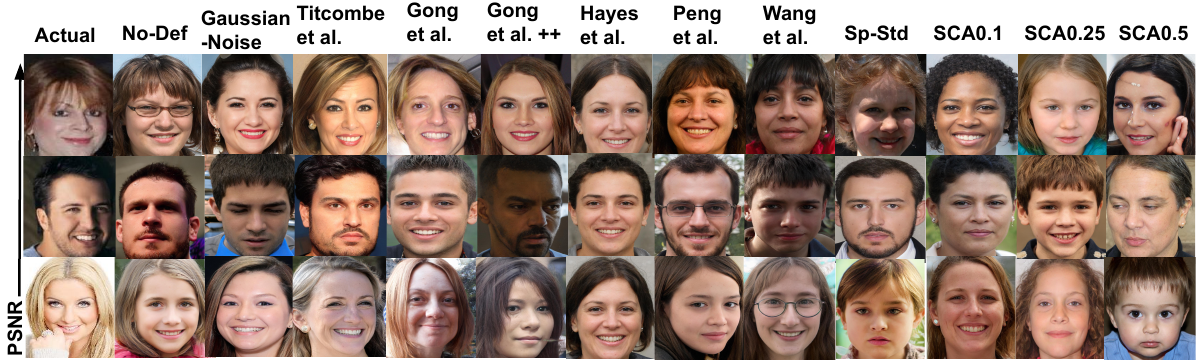}
\caption{Experiments set 1: Qualitative comparisons among actual and reconstructed images (Plug-\&-Play Attack~\cite{struppek2022plug}) under \textsc{SCA} \& baselines on hi-res CelebA dataset. 
}
\label{fig:recon_comparison_celeba}
\end{figure*}

\medskip
\noindent{\textbf{\emph{Results of experiments set 1: Plug-\&-Play attack.\\}}}
\label{ssec:resultspnp}
 \noindent \ \emph{Qualitative evaluations.}
\label{ssec:qualitative_cmp1}
Fig.~\ref{fig:recon_comparison_celeba} shows hi-res CelebA reconstructions generated by Plug-\&-Play under various defenses. To avoid cherry-picking, Fig.~\ref{fig:recon_comparison_celeba} shows the $3$ images with the highest (top row), median (middle), and lowest (bottom) PSNR reconstructions under No-Defense. Note that reconstructions under \textsc{SCA} totally differ from actual images (different race, hair, gender, child/adult), while those of other defenses are closer to actual images, indicating privacy leakage.

\noindent \ \ \ \emph{Metric evaluations.} Table~\ref{table:performance_pnp} reports reconstruction quality measures and accuracy for \textsc{SCA} and other baselines in the Plug-\&-Play attack~\cite{struppek2022plug} setting (\emph{lower rows = better defense performance}). In terms of PSNR and SSIM, training data reconstructions under the \emph{least sparse} version \mbox{\textsc{SCA0.1}} are degraded by factors of  $1.01$ to $6.7$ and $1.01$ to $5.6$ compared to the regularization defenses of \mbox{Peng et al.~\cite{peng2022bilateral}} and \mbox{Wang et al.\cite{wang2021improving}}, respectively. Increasing \textsc{SCA}'s sparsity $\lambda$ to $0.5$ widens the performance gap, increasing these factors to  $1.1$ to $6.9$ and $1.02$ to $5.9$, respectively, and making SCA outperform baselines' FID. \textsc{SCA0.1} also outperforms the noise-based approaches of \mbox{Hayes et al.\cite{hayes2023bounding}} and Titcombe et al. \cite{titcombe2021practical} on all metric by factors of $1.01$ to $5.9$ and $1.02$ to $6.3$.  Increasing \textsc{SCA}'s sparsity $\lambda$ to $0.5$ increases these factors to $1.1$ to $6.1$ and $1.1$ to $6.5$. Finally, \textsc{SCA0.1} outperforms the data augmentation defense of \cite{gong2023gan} by factors of $1.01$ to $8.4$, which widens to $1.1$ to $8.7$ for \textsc{SCA0.5}. All baselines also outperform common GAN and Noise-based industry defenses.

\smallskip
\noindent{\textbf{Basic \textsc{Sparse-Standard} outperforms SOTA baselines.}}
Our basic \textsc{Sparse-Standard} baseline outperforms the best baselines' PSNR on both CelebA and Medical MNIST, and also outperforms baselines' SSIM on the latter. \textsc{SCA0.5} then outperforms \textsc{Sparse-Standard} on \emph{all metrics}, obtaining SSIM and FID that are better by factors of $1.2$ to $1.6$ and $1.1$ to $1.2$, respectively, and slightly better PSNR. Thus, while \textsc{Sparse-Standard} offers an inferior defense vs. \textsc{SCA}, it still offers a fast practical defense for less privacy-critical domains.

\begin{table}[t]
\centering
\vspace{-1em}
\caption{Experiments set 1: Performance comparison under Plug-\&-Play attack~\cite{struppek2022plug} setting \emph{(lower rows=better defense)} on hi-res CelebA faces and Medical MNIST images.}
\begin{tabular}{ p{1.3cm} p{3.5cm} p{1.8cm} p{1.5cm} p{2.0cm} p{1.5cm}  }
\hline
 Dataset & Defense   & PSNR $\downdownarrows$    &SSIM $\downdownarrows$  &\mbox{FID  $\upuparrows$}  & Accuracy\\
\hline

   CelebA&  \textsc{No-Defense}  &$11.35$	&$0.718	$&$256.4	$&$0.779$\\
    & \textsc{Gaussian-Noise}  &$10.39	$&$0.604	$&$264.8	$&$0.644$ \\
 & \textsc{GAN}  &$10.15$	&$0.613	$&$289.7	$&$0.635$ \\
  & \mbox{Titcombe et al.\cite{titcombe2021practical}}   &$10.18	$&$0.636	$&$304.1	$&$0.654$\\
  & \mbox{Gong et al.\cite{gong2023gan}}++  &$10.02	$&$0.556	$&$381.5	$&$0.672$\\
  & \mbox{Gong et al.\cite{gong2023gan}}  &$10.11	$&$0.595	$&$350.5	$&$0.614$\\
         & \mbox{Peng et al.\cite{peng2022bilateral}}  &$9.90  $&$0.514  $&$402.8  $&$0.728$\\
&\mbox{Hayes et al.\cite{hayes2023bounding}}  &$9.92	$&$0.556	$&$383.7	$&$0.621$\\
     & \mbox{Wang et al.\cite{wang2021improving}}  &$9.85	$&$0.527	$&$402.8	$&$0.742$\\
    & \textsc{Sparse-Standard}  &$9.84	$&$0.539	$&$374.7	$&$0.728$ \\
& \textbf{\textsc{SCA0.1}} & $\mathbf{9.79}$	&$\mathbf{0.451}$	&$\mathbf{391.9}$ & $\mathbf{0.726}$ \\ 
& \textbf{\mbox{\textsc{SCA0.25}}} & $\mathbf{9.45}$	&$\mathbf{0.442}$	&$\mathbf{411.0}$ & $\mathbf{0.739}$ \\ 
& \textbf{\textsc{SCA0.5}} & $\mathbf{9.40}$	&$\mathbf{0.440}$	&$\mathbf{412.6}$ & $\mathbf{	0.723}$ \\ 
\hline
\hline
Medical & \textsc{No-Defense}  &$22.04	$&$0.396	$&$196.1	$&$0.998$\\ 
 MNIST& \textsc{Gaussian-Noise}  &$21.83$	&$0.382	$&$209.4	$&$0.862 $\\
 & \textsc{GAN}  &$21.77$	&$0.427$	&$219.0	$&$0.998 $\\
     & \mbox{Gong et al.\cite{gong2023gan}}++  &$21.50$	&$0.359$	&$273.1$	&$0.894$ \\
  & \mbox{Titcombe et al.\cite{titcombe2021practical}}  &$21.68$	&$0.360$	&$286.3$	&$0.899$ \\
    &  \mbox{Gong et al.\cite{gong2023gan}}  &$21.75$	&$0.477$	&$249.1$	&$0.770$ \\
   & \mbox{Peng et al.\cite{peng2022bilateral}}  &$21.82 $&$0.381  $&$268.3  $&$0.927$\\
& \mbox{Hayes et al.\cite{hayes2023bounding}}  &$21.72	$&$0.337	$&$259.7	$&$0.823$\\
    & \mbox{Wang et al.\cite{wang2021improving}}  &$21.71	$&$0.322	$&$211.7	$&$0.937$\\
   & \textsc{Sparse-Standard}   &$20.97$	&$0.086$	&$239.3$		&$0.907$ \\
& \textbf{\textsc{SCA0.1}} & $\mathbf{21.19}$	&$\mathbf{0.057}$	&$\mathbf{253.5}$ & $\mathbf{0.888}$ \\ 
& \textbf{\mbox{\textsc{SCA0.25}}} & $\mathbf{21.17}$	&$\mathbf{0.075}$	&$\mathbf{280.1}$ & $\mathbf{0.882}$ \\
& \textbf{\textsc{SCA0.5}} & $\mathbf{20.06}$	&$\mathbf{0.055}$	&$\mathbf{288.8}$ & $\mathbf{0.881}$ \\ 
\hline
\end{tabular}
\label{table:performance_pnp}
\vspace{-2em}
\end{table}


\begin{figure*}[t]
\centering
  \centering
  \includegraphics[width=11.2cm]
  {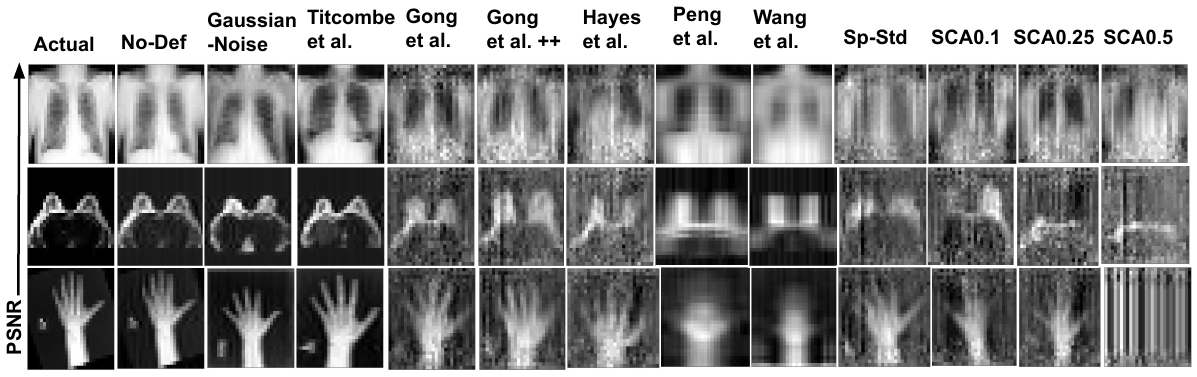}
\caption{Experiments set 2: Qualitative comparisons among actual \& reconstructed images (\emph{end-to-end} setting) under \textsc{SCA} \& baselines on the Medical MNIST dataset. 
}
\label{fig:recon_comparison_medmnist}
\end{figure*}

\begin{table}[H]
\vspace{-1em}
\centering
\caption{Experiments set 2: Performance comparison in  \emph{end-to-end} setting \emph{(lower rows=better defense)} on hi-res CelebA faces and  Medical MNIST images.}
\begin{tabular}{ p{1.3cm} p{3.5cm} p{1.8cm} p{1.5cm} p{2.0cm} p{1.5cm}  }
\hline
 Dataset & Defense   & PSNR $\downdownarrows$    &SSIM $\downdownarrows$  &\mbox{FID  $\upuparrows$}  & Accuracy\\
\hline

    CelebA&  \textsc{No-Defense}  &$16.26$	&$0.262	$&$201.8	$&$0.773$\\
    & \textsc{Gaussian-Noise}  &$16.08	$&$0.262	$&$220.4	$&$0.638$ \\
 & \textsc{GAN}  &$13.55$	&$0.133	$&$199.6	$&$0.668$ \\
  & \mbox{Titcombe et al.\cite{titcombe2021practical}}   &$15.13	$&$0.191	$&$197.7	$&$0.695$\\

  & \mbox{Gong et al.\cite{gong2023gan}}++  &$13.10	$&$0.032	$&$204.8	$&$0.704$\\
  & \mbox{Gong et al.\cite{gong2023gan}}  &$13.15	$&$0.119	$&$199.6	$&$0.682$\\
         & \mbox{Peng et al.\cite{peng2022bilateral}}  &$13.78  $&$0.141  $&$218.8  $&$0.716$\\
&\mbox{Hayes et al.\cite{hayes2023bounding}}  &$14.10	$&$0.004	$&$199.0	$&$0.664$\\
     & \mbox{Wang et al.\cite{wang2021improving}}  &$13.63	$&$0.0011	$&$203.2	$&$0.744$\\
    & \textsc{Sparse-Standard}  &$13.09	$&$0.002	$&$222.1	$&$0.749$ \\
& \textbf{\textsc{SCA0.1}} & $\mathbf{12.89}$	&$\mathbf{0.004}$	&$\mathbf{228.5}$ & $\mathbf{0.748}$ \\ 
& \textbf{\mbox{\textsc{SCA0.25}}} & $\mathbf{12.73}$	&$\mathbf{0.004}$	&$\mathbf{218.8}$ & $\mathbf{0.737}$ \\ 
& \textbf{\textsc{SCA0.5}} & $\mathbf{12.42}$	&$\mathbf{0.001}$	&$\mathbf{231.9}$ & $\mathbf{	0.741}$ \\ 

\hline
\hline

Medical& \textsc{No-Defense}  &$31.48	$&$0.935	$&$10.66	$&$0.998$\\ 
MNIST & \textsc{Gaussian-Noise}  &$30.46$	&$0.920	$&$12.23	$&$0.862 $\\
 & \textsc{GAN}  &$27.34$	&$0.480$	&$33.77	$&$0.998 $\\
     & \mbox{Gong et al.\cite{gong2023gan}}++  &$18.37$	&$0.353$	&$81.52$	&$0.894$ \\
  & \mbox{Titcombe et al.\cite{titcombe2021practical}}  &$21.33$	&$0.431$	&$30.60$	&$0.899$ \\
    &  \mbox{Gong et al.\cite{gong2023gan}}  &$21.52$	&$0.436$	&$64.88$	&$0.770$ \\
     & \mbox{Peng et al.\cite{peng2022bilateral}}  &$19.05 $&$0.420  $&$107.9  $&$0.908$\\
& \mbox{Hayes et al.\cite{hayes2023bounding}}  &$18.48	$&$0.007	$&$150.9	$&$0.824$\\
  
    & \mbox{Wang et al.\cite{wang2021improving}}  &$20.48	$&$0.549	$&$30.01	$&$0.946$\\

   & \textsc{Sparse-Standard}   &$14.79$	&$0.119$	&$250.6$		&$0.907$ \\
& \textbf{\textsc{SCA0.1}} & $\mathbf{13.43}$	&$\mathbf{0.004}$	&$\mathbf{352.1}$ & $\mathbf{0.888}$ \\
& \textbf{\mbox{\textsc{SCA0.25}}} & $\mathbf{12.32}$	&$\mathbf{0.004}$	&$\mathbf{375.9}$ & $\mathbf{0.882}$ \\ 
& \textbf{\textsc{SCA0.5}} & $\mathbf{12.04}$	&$\mathbf{0.003}$	&$\mathbf{369.9}$ & $\mathbf{0.881}$ \\ 
\hline
\end{tabular}
\label{table:performance_etn}
\vspace{-1em}
\end{table}

\paragraph{\textbf{Results of experiments set 2: End-to-end networks.\\}}\noindent \ \emph{Qualitative evaluations.}
\label{ssec:qualitative_cmp2}
Fig.~\ref{fig:recon_comparison_medmnist}, shows Medical MNIST reconstructions in the end-to-end threat model. \textsc{SCA}'s reconstructions are visually destroyed (esp. \textsc{SCA0.5}), whereas baselines admit noisy-but-recognizable reconstructions.

\label{ssec:resultsendtoend}
\noindent \ \ \ \emph{Metric evaluations.} Table~\ref{table:performance_etn}  reports performance in the end-to-end network setting. In this setting, SCA's performance advantage \emph{widens}. In terms of PSNR \& FID, training data reconstructions under \textsc{SCA0.1} are degraded by factors of $1.1$ to $3.3$ and $1.1$ to $11.7$ compared to  \mbox{Peng et al.~\cite{peng2022bilateral}} and \mbox{Wang et al.\cite{wang2021improving}}, respectively (but slightly worse SSIM vs. Wang et al. on CelebA). On all metrics, \textsc{SCA0.1} outperforms \mbox{Hayes et al.\cite{hayes2023bounding}} and Titcombe et al. \cite{titcombe2021practical} by factors of $1.1$ to $2.4$ and $1.1$ to $107.7$, respectively. \textsc{SCA0.1} also outperforms Gong et al. \cite{gong2023gan} by factors of $1.01$ to $109$. Increasing \textsc{SCA}'s $\lambda$ to $0.5$ causes it to outperform the same baselines \emph{on all metrics} by  factors of $1.2$ to $141$ \cite{peng2022bilateral},  $1.2$ to $183$ \cite{wang2021improving}, $1.2$ to $4.0$\cite{hayes2023bounding}, $1.2$ to $191$ \cite{titcombe2021practical}, and $1.1$ to $145.3$ \cite{gong2023gan}, respectively.

\begin{figure}[h]

\begin{subfigure}{.5\textwidth}
  \centering
    \includegraphics[width=5.4cm, height=3.0cm]{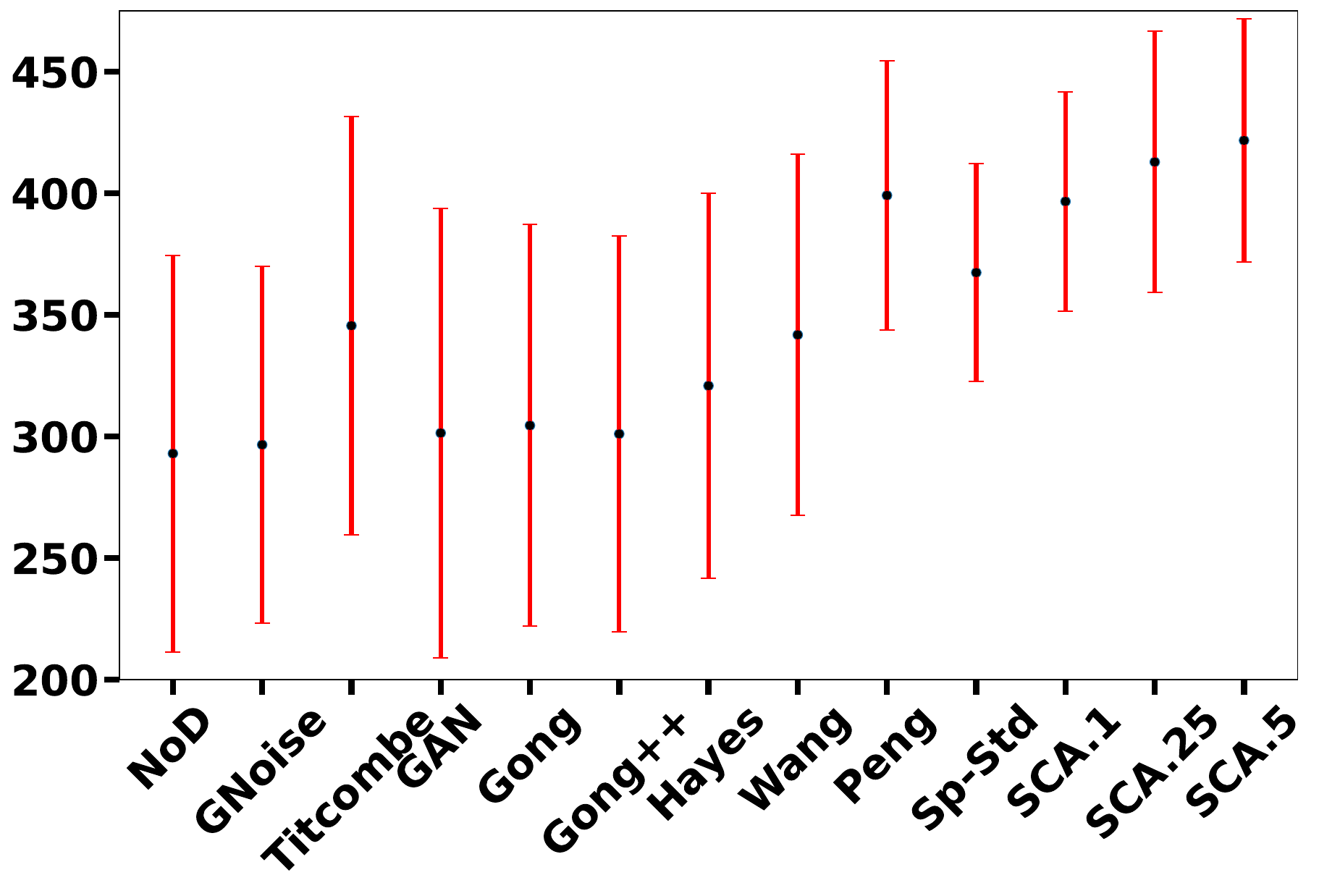}
  \caption{FID$\upuparrows$ for CelebA faces under Plug-\&-Play}
  \label{fig:dist_FID_celeba}
\end{subfigure}
\begin{subfigure}{.5\textwidth}
  \centering
    \includegraphics[width=5.4cm, height=3.0cm]{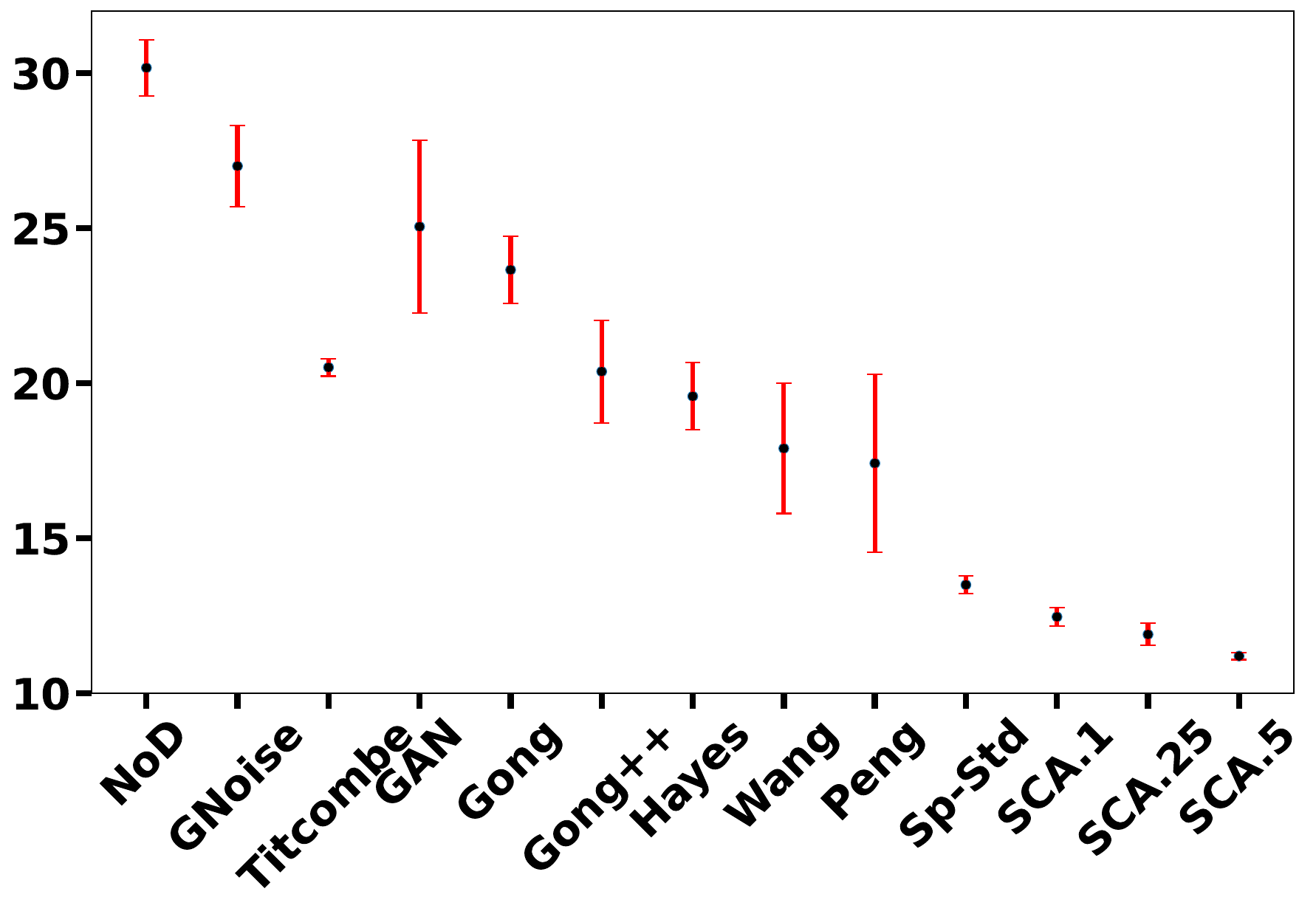}
  \caption{PSNR$\downdownarrows$ for Med.MNIST, end-to-end setting}
  \label{fig:dist_psnr_mmnist}
\end{subfigure}%
\caption{Stability of SCA \& baselines' defense performance (mean $\pm$ std. dev.) of PSNR and FID across multiple runs on CelebA and Medical MNIST.}
\label{fig:bar_plot_dist}
\end{figure}

\paragraph{\textbf{Stability of \textsc{SCA}'s defense vs. baselines.}}
\label{ssec:results_stability}
\textsc{SCA}'s performance is also as stable or more stable than baselines' performance over multiple runs. For example, 
Fig. \ref{fig:dist_FID_celeba} plots the means \& std. devs. of SCA \& baselines' per-run FID (the standard metric for faces) over multiple runs for CelebA faces in the Plug-\&-Play setting, and Fig. \ref{fig:dist_psnr_mmnist} plots this for PSNR (the standard metric for grayscale images) on Medical MNIST in the end-to-end setting. SCA obtains better performance while also exhibiting stability of defense performance on par with the best baseline (and better stability than most baselines). See Appendix E.

\begin{table}[h]
\centering
\caption{Experiments set 3: Performance comparison in \emph{split network} setting \emph{(lower rows=better defense)} on hi-res CelebA faces and sensitive Medical MNIST images.}
\begin{tabular}{ p{1.3cm} p{3.5cm} p{1.8cm} p{1.5cm} p{2.0cm} p{1.5cm}  }
\hline
 Dataset & Defense   & PSNR $\downdownarrows$    &SSIM $\downdownarrows$  &\mbox{FID  $\upuparrows$}  & Accuracy\\
\hline

    CelebA&  \textsc{No-Defense}  &$16.49$	&$0.302	$&$185.8	$&$0.766$\\
    & \textsc{Gaussian-Noise}  &$15.44	$&$0.227	$&$191.1	$&$0.753$ \\
 & \textsc{GAN}  &$15.57$	&$0.253	$&$176.7	$&$0.646$ \\
  & \mbox{Titcombe et al.\cite{titcombe2021practical}}   &$14.99	$&$0.144	$&$194.2	$&$0.725$\\

  & \mbox{Gong et al.\cite{gong2023gan}}++  &$15.06	$&$0.038	$&$190.5	$&$0.756$\\
  & \mbox{Gong et al.\cite{gong2023gan}}  &$15.65	$&$0.044	$&$185.8	$&$0.653$\\
         & \mbox{Peng et al.\cite{peng2022bilateral}}  &$16.23  $&$0.211  $&$198.6  $&$0.717$\\

&\mbox{Hayes et al.\cite{hayes2023bounding}}  &$15.06	$&$0.005	$&$178.8	$&$0.672$\\
     & \mbox{Wang et al.\cite{wang2021improving}}  &$14.82	$&$0.173	$&$189.6	$&$0.652$\\
    & \textsc{Sparse-Standard}  &$15.39	$&$0.009	$&$187.0	$&$0.746$ \\
& \textbf{\textsc{SCA0.1}} & $\mathbf{15.05}$	&$\mathbf{0.005}$	&$\mathbf{178.7}$ & $\mathbf{0.745}$ \\ 
& \textbf{\mbox{\textsc{SCA0.25}}} & $\mathbf{14.76}$	&$\mathbf{0.003}$	&$\mathbf{191.1}$ & $\mathbf{0.743}$ \\ 
& \textbf{\textsc{SCA0.5}} & $\mathbf{14.71}$	&$\mathbf{0.003}$	&$\mathbf{206.1}$ & $\mathbf{	0.739}$ \\ 

\hline
\hline
Medical& \textsc{No-Defense}  &$23.47	$&$0.776	$&$45.57	$&$0.993 $\\ 
MNIST & \textsc{Gaussian-Noise}  &$21.93$	&$0.722	$&$44.72	$&$0.811 $\\
 & \textsc{GAN}  &$21.67$	&$0.719$	&$48.49	$&$0.912 $\\
     & \mbox{Gong et al.\cite{gong2023gan}}++  &$21.07$	&$0.573$	&$67.53$	&$0.931$ \\
  & \mbox{Titcombe et al.\cite{titcombe2021practical}}  &$21.35$	&$0.704$	&$48.82$	&$0.961$ \\
    &  \mbox{Gong et al.\cite{gong2023gan}}  &$21.33$	&$0.720$	&$41.74$	&$0.925$ \\

 & \mbox{Peng et al.\cite{peng2022bilateral}}  &$18.98 $&$0.426  $&$124.8  $&$0.914$\\
 &   \mbox{Hayes et al.\cite{hayes2023bounding}}  &$21.46	$&$0.442	$&$137.4	$&$0.850$\\
    & \mbox{Wang et al.\cite{wang2021improving}}  &$20.03	$&$0.538	$&$65.17	$&$0.986$\\
   & \textsc{Sparse-Standard}   &$15.33$	&$0.149$	&$142.4$		&$0.955$ \\
& \textbf{\textsc{SCA0.1}} & $\mathbf{13.95}$	&$\mathbf{0.008}$	&$\mathbf{244.9}$ & $\mathbf{0.946}$ \\ 
& \textbf{\mbox{\textsc{SCA0.25}}} & $\mathbf{12.31}$	&$\mathbf{0.008}$	&$\mathbf{255.3}$ & $\mathbf{0.928}$ \\ 
& \textbf{\textsc{SCA0.5}} & $\mathbf{12.27}$	&$\mathbf{0.001}$	&$\mathbf{285.3}$ & $\mathbf{0.909}$ \\ 
\hline
\end{tabular}

\label{table:performance_split}
\end{table}

\paragraph{\textbf{\textsc{SCA}'s sparsity vs. performance.}}
\label{ssec:results_lambda}
We also try varying \textsc{SCA}'s and \textsc{Sparse-Standard}'s sparsity parameters $\lambda$ and recompute PSNR, SSIM, FID, and accuracy.
Appendix A.6 shows that for each $\lambda$ and defense metric, $\textsc{SCA}$ significantly outperforms the off-the-shelf \textsc{Sparse-Standard} architecture for a small accuracy cost. Thus, for a given $\lambda$ with \textsc{Sparse-Standard}, we can use a (smaller) $\lambda$ with \textsc{SCA} to obtain better reconstruction \emph{and} higher or equal (within $0.0017$) accuracy. \textsc{SCA} is also amenable to more sophisticated tuning (and performance improvements) by tuning different $\lambda$ per sparse layer (e.g., by having a sparser representation of inputs but less sparse reductions of downstream layers). We \emph{avoid} such tuning here as it is unnecessary for good performance.

\paragraph{\textbf{Results of experiments set 3: Split networks.}}
\label{ssec:resultssplit}
Table~\ref{table:performance_split} reports performance in the split network setting. As expected, all baselines and SCA perform slightly worse in this threat model compared to the end-to-end model (aside from Guassian and GAN heuristics). On all metrics, SCA's performance advantage remains consistent: \textsc{SCA0.5} outperforms all baselines by factors of \mbox{$1.1$ to $720$}.

\section{Empirical analysis of sparse coding robustness to attack}\label{sec:umaps}

\begin{figure}[t]
\begin{subfigure}{.33\textwidth}
  \centering
  \includegraphics[width=3.0cm]{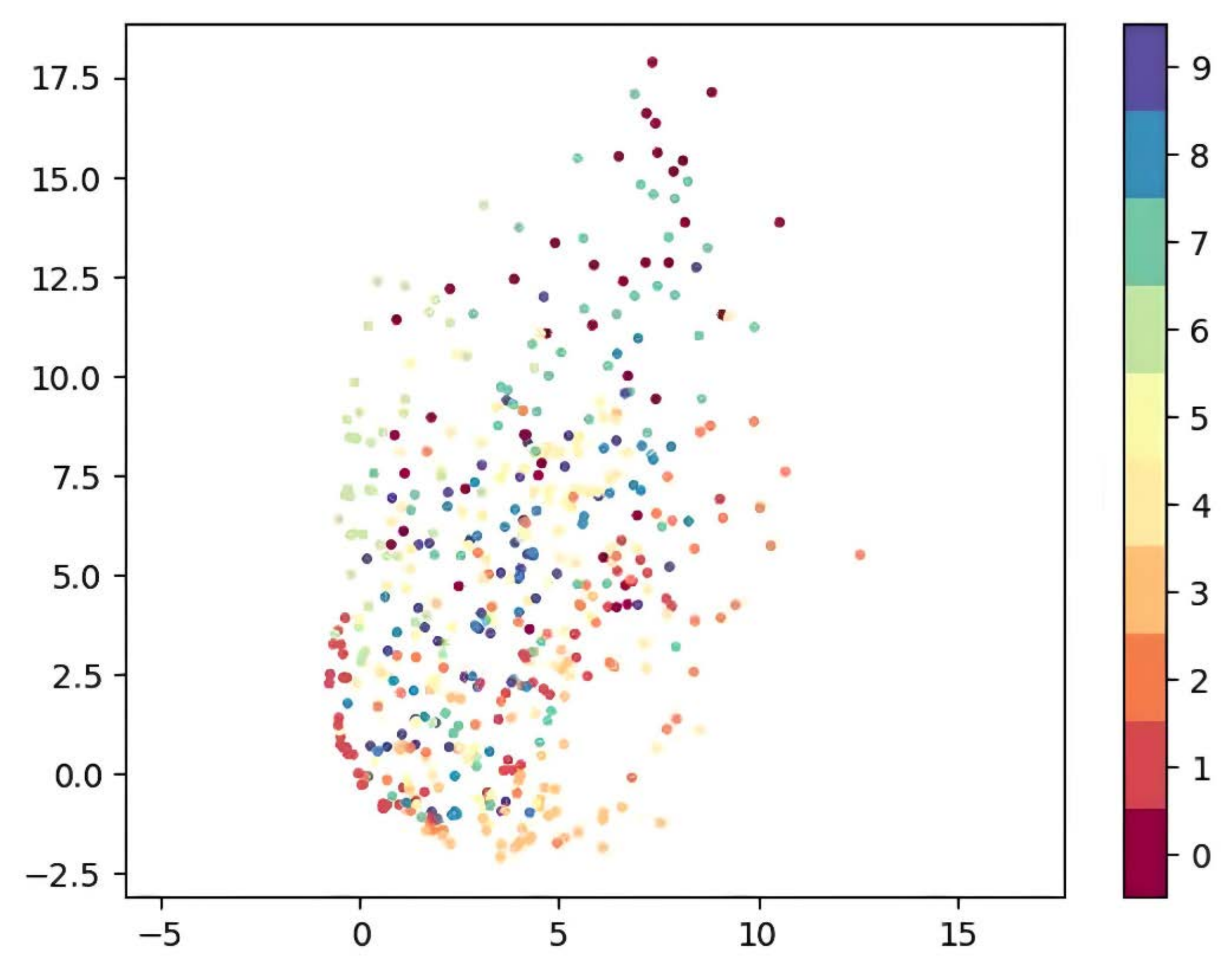}
  \caption{Linear Layer}
  \label{fig:dist_linear_one}
\end{subfigure}%
\begin{subfigure}{.33\textwidth}
  \centering
  \includegraphics[width=3.0cm]{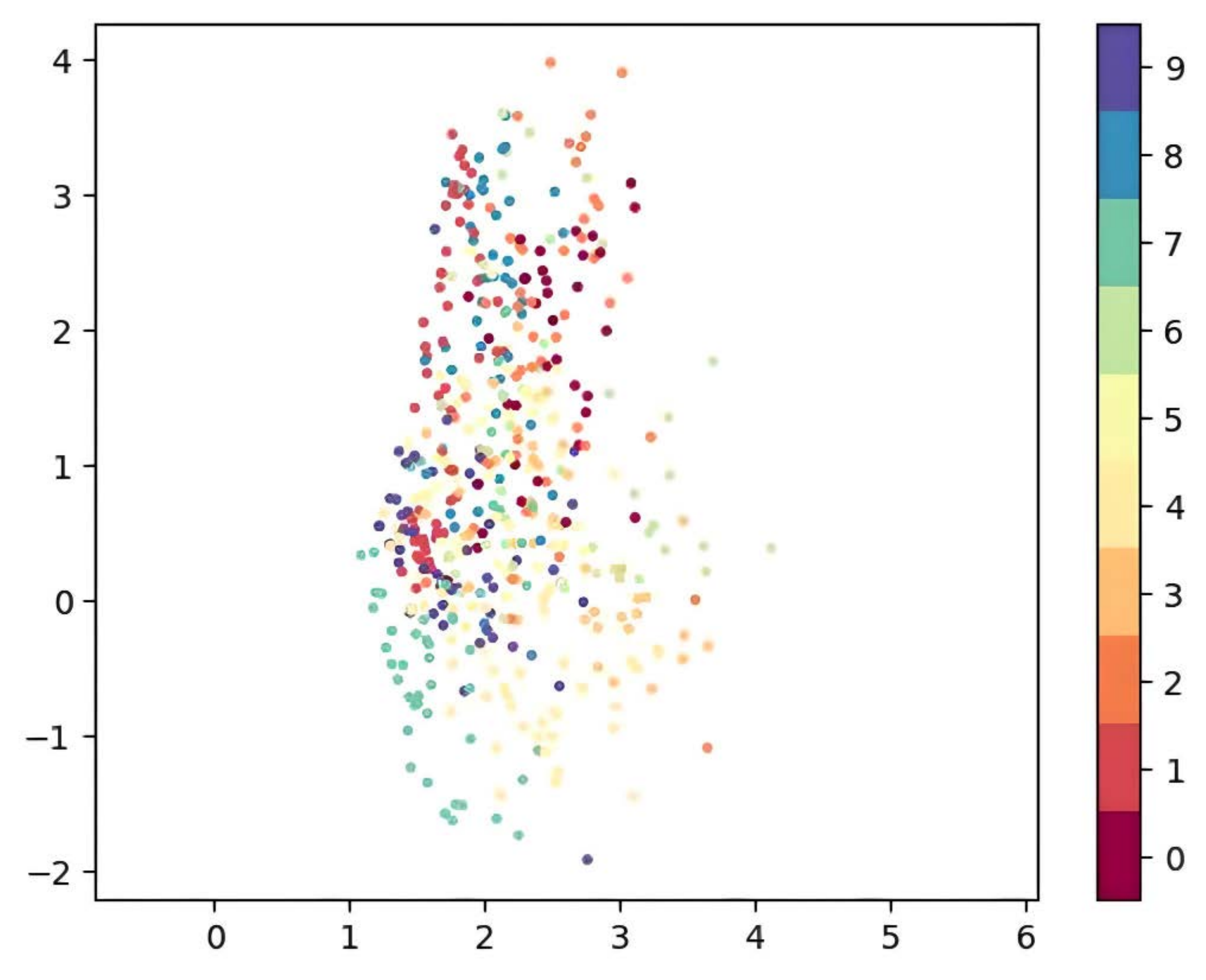}
  \caption{Convolution Layer}
  \label{fig:dist_cnn_one}
\end{subfigure}
\begin{subfigure}{.33\textwidth}
  \centering
  \includegraphics[width=3.0cm]{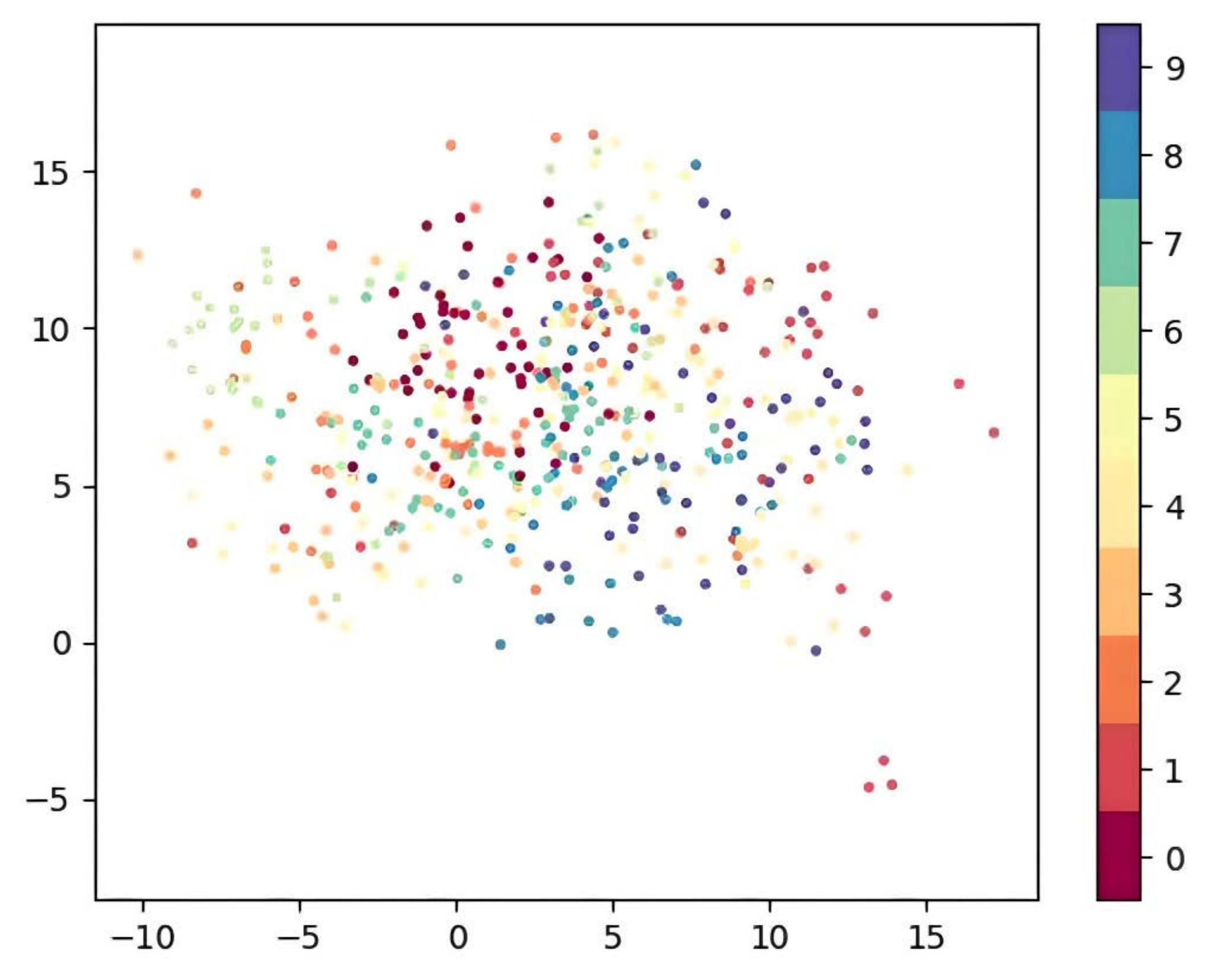}
  \caption{Sparse Coding Layer}
  \label{fig:dist_lca_one}
\end{subfigure}
\begin{subfigure}{.33\textwidth}
  \centering
  \includegraphics[width=3.0cm]{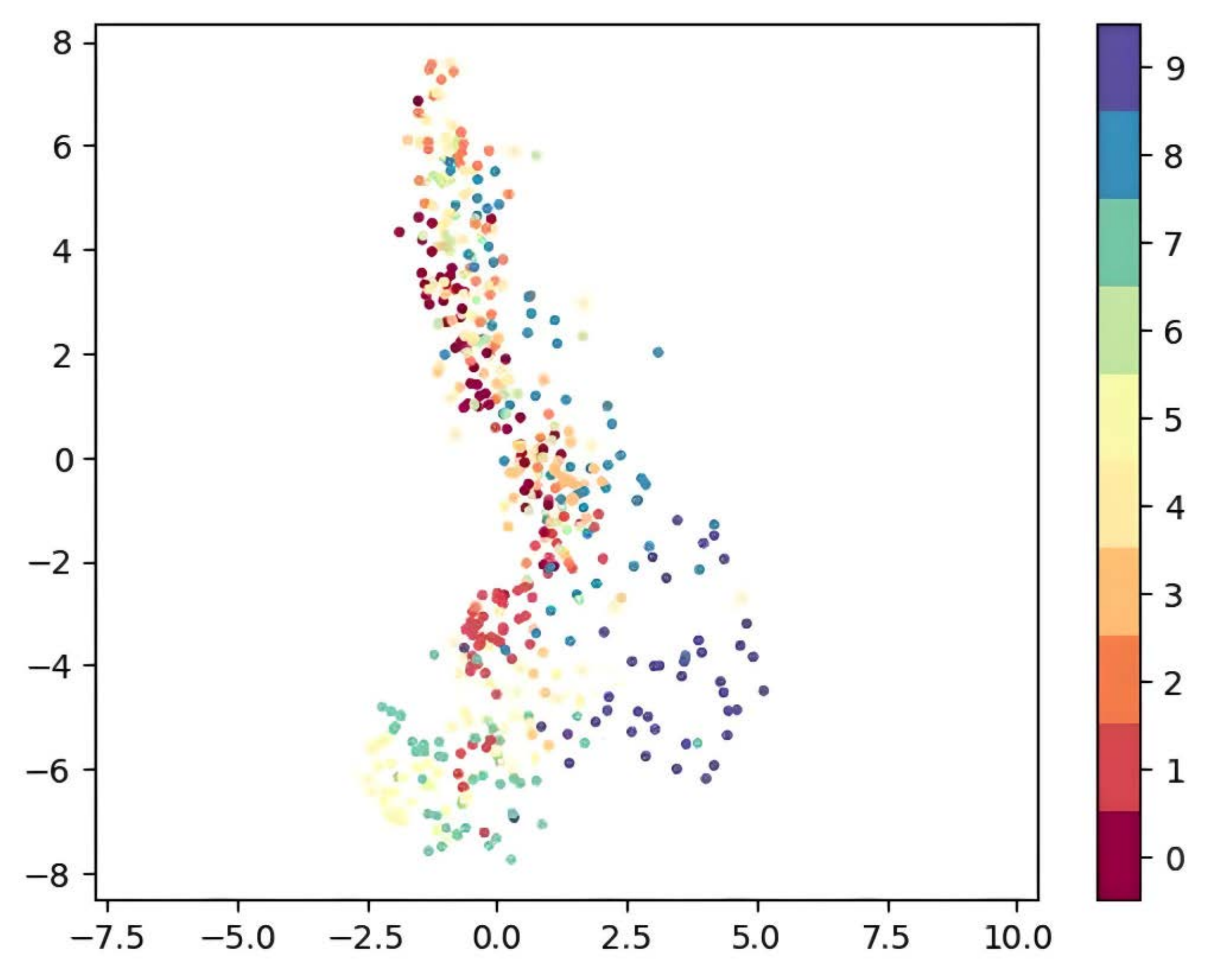}
  \caption{Linear Layer}
  \label{fig:dist_linear}
\end{subfigure}%
\begin{subfigure}{.33\textwidth}
  \centering
  \includegraphics[width=3.0cm]{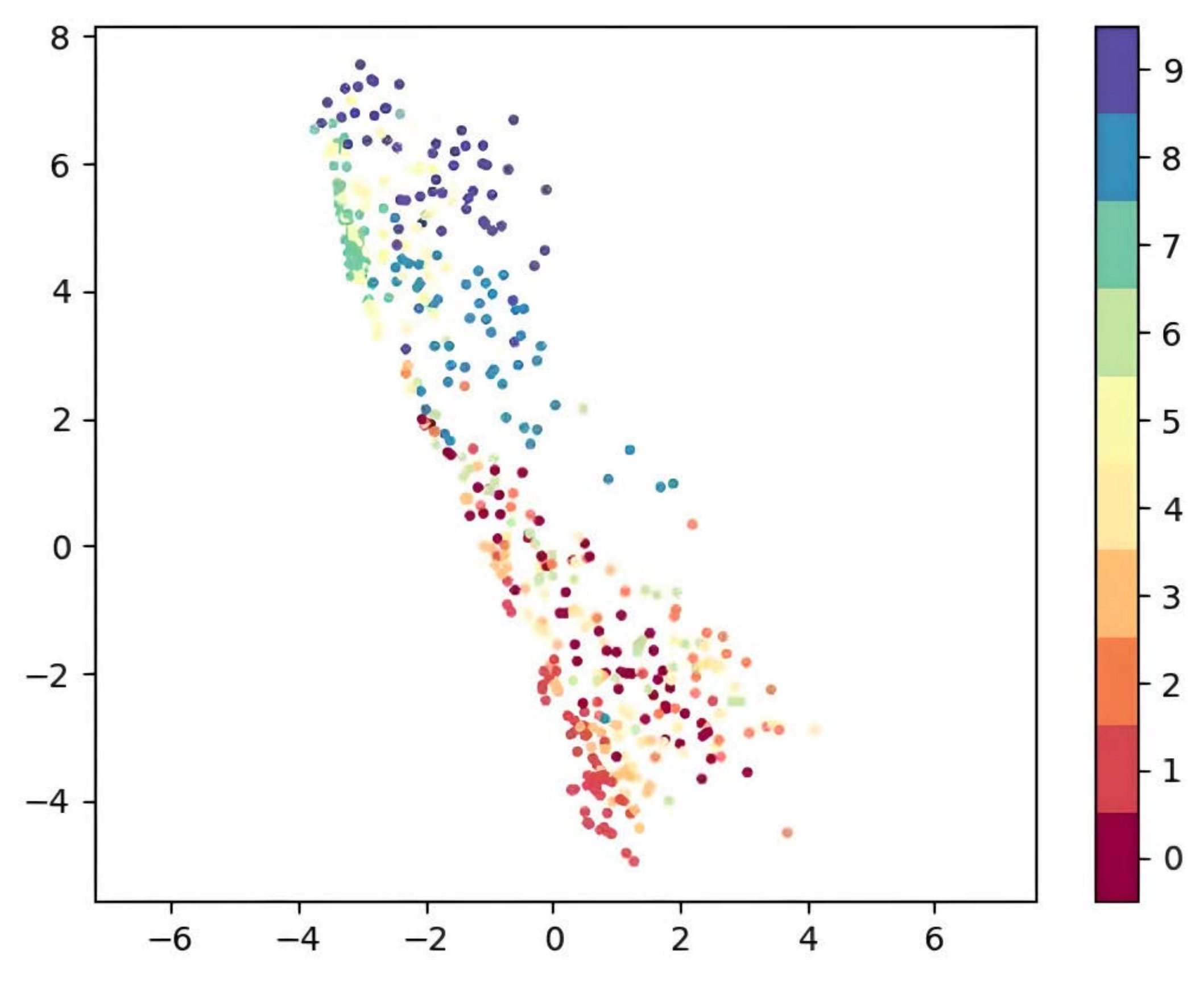}
  \caption{Convolution Layer}
  \label{fig:dist_cnn}
\end{subfigure}
\begin{subfigure}{.33\textwidth}
  \centering
  \includegraphics[width=3.0cm]{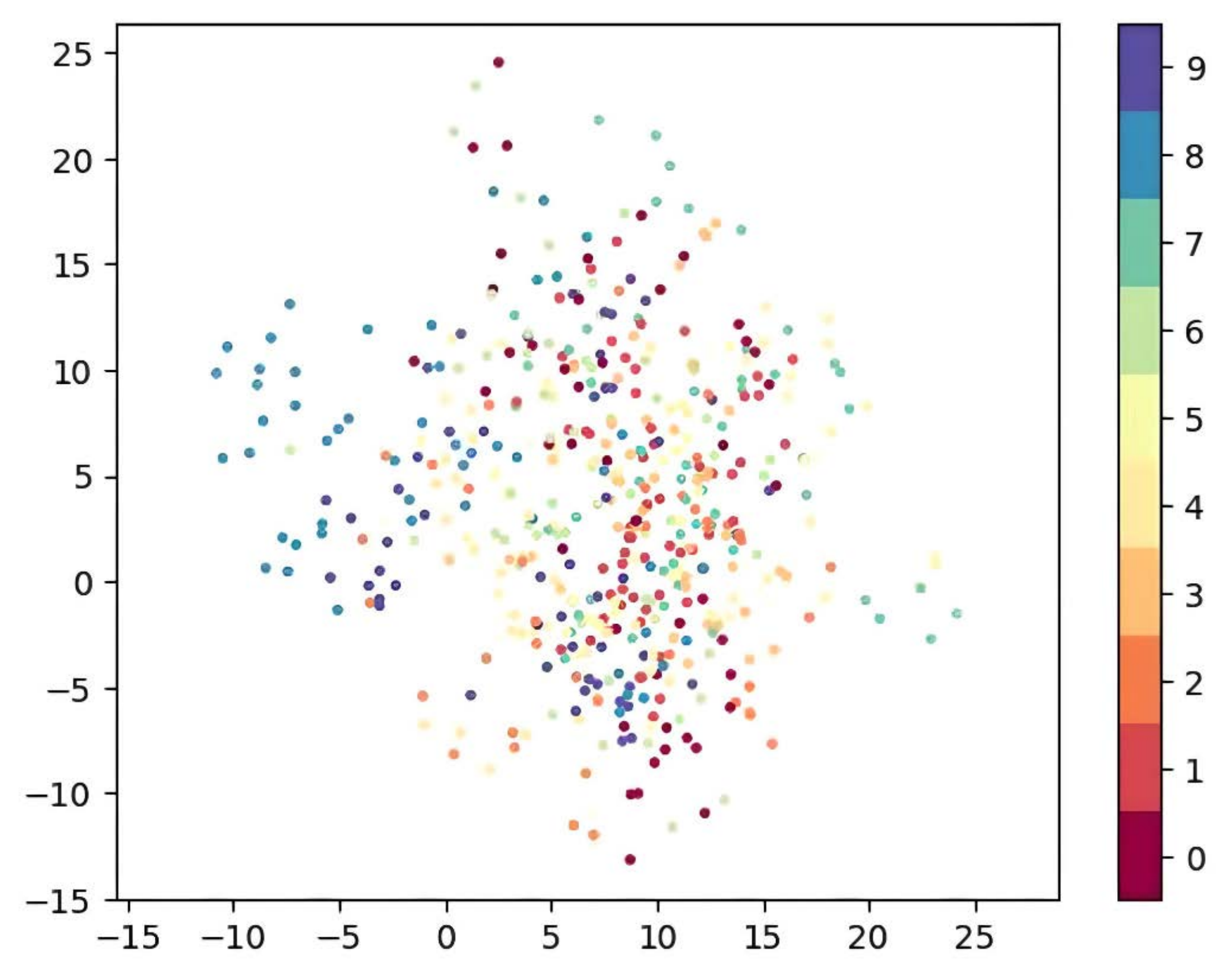}
  \caption{Sparse Coding Layer}
  \label{fig:dist_lca}
\end{subfigure}
\caption{ UMap 2D projections of input images' features by class after $2$ linear layers, $2$ conv. layers, or $2$ sparse-coded layers on MNIST (top) \& Fashion MNIST (bottom).}
\label{fig:dist_umap}
\end{figure}

Sparse-coding layers' robustness to privacy attacks can be observed empirically. Consider that the attacker trains the attack to map leaked raw hidden layer outputs back to input images. Attacks are thus highly dependent on these outputs' distributions. Recall that UMAP projections compute a 2D visualization of the global structure of distances between different training images' features according to a particular layer \cite{mcinnes2018umap}. Fig. \ref{fig:dist_umap} plots UMAP 2D projections of linear layer feature distributions of training inputs \emph{after} either two linear layers (Figs. \ref{fig:dist_linear_one} \& \ref{fig:dist_linear}), two convolutional layers (Figs. \ref{fig:dist_cnn_one} \& \ref{fig:dist_cnn}), or two sparse coding layers (with interspersed dense layers -- Figs. \ref{fig:dist_lca_one} \& \ref{fig:dist_lca}). Importantly, observe that after two linear or two convolutional layers, points are clustered by color, i.e., input images' features are highly clustered by label. This class-clustered property leaves such layers vulnerable to model inversion attacks, as an attacker can `home in on' examples from a specific class. In contrast, the goal in sparse coding is not to optimize the classification objective by separating classes, but rather to jettison unnecessary information. Here, this means that unnecessary information is jettisoned both from the input image and also the downstream dense layer. Per Figs. \ref{fig:dist_lca_one} \& \ref{fig:dist_lca}, this tends to `uncluster' remaining non-sparsified features of training examples from the same class, making it much harder for an attacker to compute informative gradients to home in on a training example.

\section{Discussion \& Conclusion}
\label{sec:conclusion}
In this paper, we have provided the first study of sparse coding-based neural network architectures that are robust to model inversion attacks. Specifically, we have shown that the natural properties of sparse coded layers can control the extraneous private information about the training data that is encoded in a network without resorting to complex and computationally intensive parameter tuning techniques. Our work reveals a deep connection between state-of-the-art privacy vulnerabilities and three decades of computer science research on sparse coding for other application domains. Currently, our basic research implementation of \textsc{SCA} achieves compute times that in the worst-case are no better than some SOTA baselines (see Appendix F). However, given the rich theoretic body of work on fast algorithms and provable guarantees for sparse coding, we believe these aspects are opportune areas for future improvements.


%
%
\section*{Acknowledgements}
We are grateful for generous support from OpenAI, as well as the Dartmouth College Cybersecurity Cluster Research. This work was partially funded by the Center for Nonlinear Studies and the Information Science and Technology Institute's Cyber Security Summer School at Los Alamos National Laboratory, as well as an award from the Department of Energy's Advanced Scientific Computing Research program (\#77902). 

\bibliographystyle{splncs04}
\bibliography{refs}

\appendix
\section{Appendix}
\label{sec:app}
This is the supplementary document containing the additional results and details of our proposed Sparse Coding Architecture (SCA) formulations, as well as cluster details and additional preliminaries.

\subsection{Reproducibility}
\label{sec:reproducibility}
In order to promote further research and standardize the evaluations of new defenses, we provide full cluster-ready PyTorch \cite{paszke2019pytorch} implementations of \textsc{SCA} and all benchmarks as well as replication codes for all experiments on our project page at: \href{https://sayantondibbo.github.io/SCA}{https://sayantondibbo.github.io/SCA}. 

We provide full details of the cluster hardware and all parameter choices used in our experiments in \textit{Appendix} \ref{ssec:cluster} and \ref{ssec:spguard_archi}, and in \textit{Appendix} Tables \ref{wrap-tab:cluster_details} and \ref{wrap-tab:details_archi_sg}.

\subsection{Adapting Rozell LCA to Convolutional Networks}\label{ssec:lcaconv}
    Although the original LCA formulation \cite{rozell2008sparse} was introduced for the non-convolutional case, it is based on the general principle of feature-similarity-based competition between neurons within the same layer, which can be  adapted to the convolutional setting via only two minimal changes to Equation \ref{eqn:ode_potential} \cite{teti2022lcanetst,kim2020modeling}. In Rozell's original formulation, $\Psi(t)$ can simply be recast from a matrix multiplication to a convolution between the input and dictionary. Second, the lateral interaction tensor, $\mathcal G$ in Equation \ref{eqn:ode_potential}, can also be recast from a matrix multiplication to a convolution between the dictionary and its transpose. Neuron membrane potential works as follows: 
\begin{equation}\label{eqn:ode_potential}
\mathcal {\dot P}(t) = \frac{1}{\tau}[\Psi (t) - \mathcal {P}(t) - \mathcal R_x(t) * \mathcal G]
\end{equation}
where $\tau$ is a time constant, $\Psi (t) = \mathcal X * \Omega$ is the neuron's bottom-up drive from the input computed by taking the convolution, $*$, between the input, $\mathcal X$, and the dictionary, $\Omega$, and $- \mathcal P (t)$ is the leak term \cite{teti2022lcanetst,kim2020modeling}.
\subsection{Cluster Details}
\label{ssec:cluster}
We run all our experiments using the slurm batch jobs on industry-standard high-performance GPU clusters with $40$ cores and $4$ nodes. Details of the hardware and architecture of our cluster are described in Table~\ref{wrap-tab:cluster_details}. We note that noise-based \textsc{Gaussian} and Titcombe et al. \cite{titcombe2021practical} defenses are typically fastest on this architecture (though they are the least-performant). We emphasize that our sparse coding implementations are `research-grade', unlike the optimized torch GAN implementations available for \cite{gong2023gan}. See also \textit{Appendix} \ref{ssec:computetime}. Note that for large scale applications, \textsc{SCA}'s sparse coding updates can be accelerated such that they can be computed extremely efficiently (see the training complexity discussion in the main paper body).

\begin{table}[h]
\centering
 \caption{Hardware Details of the Cluster in our Experiments.}
 \label{wrap-tab:cluster_details}
\begin{tabular}{cc}\\\toprule  
 $Parameter$  & \textsc{Measurements}   \\ \midrule
 Core &40	\\  \midrule
RAM &565GB	\\  \midrule
GPU &Tesla V100	\\ \midrule 
  Nodes &p01-p04	\\  \midrule

Space &1.5TB	\\  

   \bottomrule
 \end{tabular}
 \end{table}

\subsection{Parameters and architecture of \textsc{SCA}}
\label{ssec:spguard_archi}
We implement \textsc{SCA} using two Sparse Coding Layers (SCL): One following the input image, and one following a downstream dense batch normalization layer. Finally, we follow these two pairs of dense-then-sparse layers with downstream fully connected (linear) layers before the classification layer. In the case of end-to-end network experiments, we use $5$ downstream linear layers, which is a reasonable default. In the split network setting, we are careful to use $3$ downstream fully connected layers in order to match the architectures used in the split network experimental setup of \cite{titcombe2021practical}, and per our public codebase, we make every effort to make the benchmarks within each setting comparable in terms of architecture, aside from the obvious difference of \textsc{SCA}'s sparse layers We train \textsc{SCA}'s sparse layers with $500$ iterations of lateral competitions during reconstructions in SCL layers. We emphasize that \textsc{SCA} can be made significantly more complex, either via the addition of more sparse-dense pairs of layers, or by adding additional (convolutional, linear) downstream layers before classification. We avoid such complexity in the experiments in order to compare more directly to benchmarks and because our goal is to study an architecture that captures the essence of \textsc{SCA}. We give all parameter and training details in Table~\ref{wrap-tab:details_archi_sg}.

\begin{table}[h]
\centering
 \caption{Architecture and Parameters of \textsc{SCA} implementation.}
 \label{wrap-tab:details_archi_sg}
\begin{tabular}{cc}\\\toprule  
 $Parameter$  & \textsc{Value}   \\ \midrule
 Sparse Layers &2	\\  \midrule
Batch Norm Layers &2	\\ \midrule 
Fully Connected Layers &5	\\  \midrule
  $\lambda$ &0.5	\\  \midrule
Learning rate $\eta$ &0.01	\\  \midrule
Time constant $\tau$ &1000	\\  \midrule
Kernel size &5\\ \midrule
Stride & 1,1\\ \midrule
Lateral competition iterations &500	\\  

   \bottomrule
 \end{tabular}
 \end{table} 

\subsection{Attack details}
\label{ssec:mi_attackexperiments}
  In the Plug-\&-Play attack experiments, we follow the authors' attack exactly ~\cite{struppek2022plug}, except we update their approach to use the latest \textbf{StyleGAN3}~\cite{Karras2021} for high-resolution image generation. For the end-to-end and split-network attacks, we consider a recent state-of-the-art surrogate model training attack optimized via SGD ~\cite{xu2023sparse,aivodji2019gamin}. This attack works by querying the target model with an externally obtained dataset. To capture a well-informed `worst-case' attacker, we set this dataset to a holdout set from the true training dataset. The attack then uses the corresponding model high-dimensional intermediate outputs to train an inverted surrogate model that outputs actual training data.

\subsection{SCA sparsity vs. robustness} We vary the sparsity, i.e., $\lambda$ parameter and run the \textsc{Sparse-Standard}, as well as our \textsc{SCA}. We observe that increasing  $\lambda$ helps improve the robustness, without significant accuracy drops. For example, Table ~\ref{tab:tuninglambda} shows this comparison for MNIST in the end-to-end setting.
\label{appendixsparserobust}

\begin{table}[h]
  \caption{\textsc{Sparse-Standard} and \textsc{SCA} performance with  $\lambda$$\in$$\{0.1, 0.25$, $0.5, 0.75\}$}
\makebox[\textwidth][c]{
    \begin{tabular}{ccccccccc}
\toprule
      &   \multicolumn{2}{c}{PSNR$\downdownarrows$}    & \multicolumn{2}{c}{SSIM$\downdownarrows$} & \multicolumn{2}{c}{FID $\upuparrows$} & \multicolumn{2}{c}{Accuracy} \\
\cmidrule(lr){2-3}\cmidrule(lr){4-5}\cmidrule(lr){6-7}\cmidrule(lr){8-9}
$\lambda$ & {\small\textsc{Sp-Std}} & {\small\textsc{SCA}}  & {\small\textsc{Sp-Std}} & {\small\textsc{SCA}}  & {\small\textsc{Sp-Std}}  & {\small\textsc{SCA}} & {\small\textsc{Sp-Std}} & {\small\textsc{SCA}}\\ \midrule 
$0.1 $&$ 23.45 $&$ 19.54 $&$ 0.650 $&$ 0.502 $&$ 111.5 $&$ 178.5 $&$  0.984    $&$ 0.984 $   \\
$0.25 $&$ 21.34 $&$ 18.81 $&$ 0.438 $&$ 0.340 $&$ 142.9 $&$ 174.1 $&$ 0.986    $&$ 0.983 $   \\
$0.5  $&$ 22.16 $&$ 17.85 $&$ 0.598 $&$ 0.164 $&$ 136.9 $&$ \mathbf{335.4} $&$ 0.985    $&$ 0.977 $   \\
$0.75 $&$ 22.39 $&$ \mathbf{14.65} $&$ 0.593 $&$ \mathbf{0.086} $&$ 142.0 $&$ 214.1 $&$ 0.981    $&$ 0.971 $  \\
\bottomrule
\end{tabular}%
}
\label{tab:tuninglambda}%
\end{table}%

\section{Model Inversion Attack Methodology: Additional discussion}
\label{ssec:mi_attack}
Because privacy attacks are an emerging field, we feel it is relevant to include additional context and discussion here. Recent work has highlighted a variety of attack vectors targeting sensitive training data of machine learning models~\cite{liu2022ml,dibbo2023model,vhaduri2021hiauth,tramer2022truth,shokri2017membership,zhang2020secret,choquette2021label,dibbo2023sok, dibbo2023model, vhaduri2022predicting,sablayrolles2019white,gong2016you,zhong2022understanding,carlini2023extracting,vhaduri2023bag,li2023sok,carlini2021extracting}. These attacks not only target centralized models but also can make the federated learning models vulnerable to attacks~\cite{he2019model, fang2023gifd}. Adversaries with different access (i.e., black-box, white-box) to these models perform different attacks leveraging a wide range of capabilities, e.g., knowledge about the target model confusion matrix and access to blurred images of that particular class~\cite{fredrikson2015model,choquette2021label,he2019model,wang2021variational,juuti2019prada}. Such attacks commonly fall under the umbrella of privacy attacks, which include specific attacker goals such as membership inference, model stealing, model inversion, etc.~\cite{mehnaz2022your,wang2022dualcf,yuan2022attack,hu2022membership}. Defending against privacy attacks is a core task of mainstream technology platforms ranging from public social networks to private medical research~\cite{breuer2023preemptive, naveed2015privacy, xu2021deep}.

\begin{figure}[t]
\begin{subfigure}{.33\textwidth}
  \centering
  \includegraphics[width=3.0cm, height=3cm]{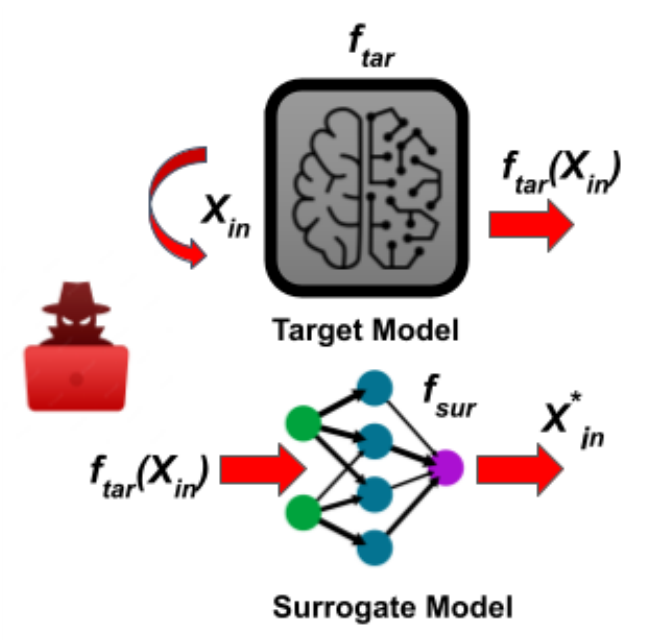}
  \caption{Model Inversion Attack}
  \label{fig:MI_overview}
\end{subfigure}%
\begin{subfigure}{.33\textwidth}
  \centering
  \includegraphics[width=4.0cm, height=2.5cm]{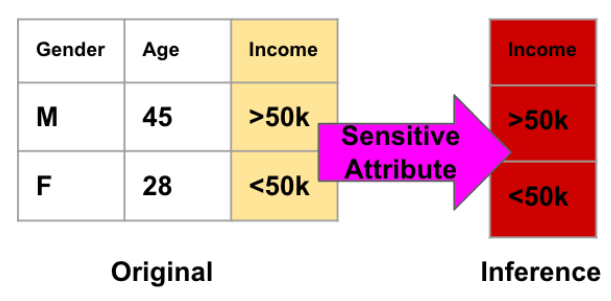}
  \caption{Attribute Inference}
  \label{fig:AI_Overview}
\end{subfigure}
\begin{subfigure}{.33\textwidth}
  \centering
  \includegraphics[width=4cm, height=2.0cm]{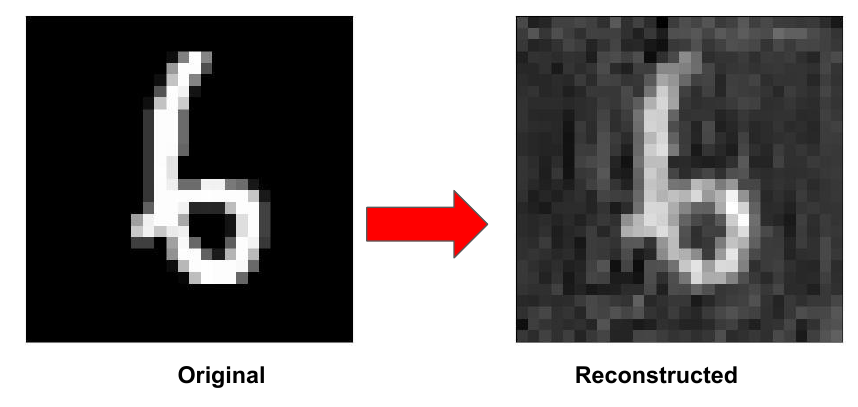}
  \caption{Image Reconstruction}
  \label{fig:IR_Overview}
\end{subfigure}
\caption{Illustration of Model Inversion attack along with (a.) pipelines--an adversary queries target model $f_{tar}$ with inputs $\mathcal X_{in}$ to obtain output $f_{tar}(X_{in})$. Then adversary trains a surrogate attack model $f_{sar}$, where the  $f_{tar}(X_{in})$ is the input and $\mathcal X^*$ is the output;
and (b.) categories, i.e., attribute inference (AttrInf) attack, where the adversary infers sensitive attribute $\mathcal X_s$ with or without knowing non-sensitive attribute values, i.e., $\mathcal X_{ns}\rightarrow \mathcal X_s$  and (c.) image reconstruction (ImRec) attack, where adversary reconstructs similar to original images, i.e., $\mathcal X_{in} \approx \mathcal X^*_{in}$.}

\label{fig:MIAI_attacks}
\end{figure}

Our focus is model inversion attack, where an adversary aims to infer sensitive training data attributes $X_s$ or reconstruct training samples $X_{in}$, a severe threat to the privacy of training data $D_{Tr}$~\cite{titcombe2021practical,mehnaz2022your}. In Figure~\ref{fig:MI_overview}, we present the pipelines of the model inversion attack. Depending on data types and purpose, model inversion attacks can be divided into two broader categories: (i) attribute inference (AttrInf) and (ii) image reconstruction (ImRec) attacks~\cite{dibbo2023sok}. In AttrInf attacks, it is assumed the adversary can query the target model $f_{tar}$ and design a surrogate model $f_{sur}$ to infer some sensitive attributes $X_s$ in training data $D_{Tr}$, with or without knowing all other non-sensitive attributes training data $X_{ns}$ in the training data $D_{Tr}$, as presented in Figure~\ref{fig:AI_Overview}. In ImRec attacks the adversary reconstructs entire training samples $D_{Tr}$ using the surrogate model $f_{sur}$ with or without having access to additional information like blurred, masked, or noisy training samples $D_{s}$, as shown in Figure~\ref{fig:IR_Overview}~\cite{fredrikson2015model,zhang2020secret,zhao2021exploiting}. To contextualize our \textsc{SCA} setting, recall that we suppose the attacker has only black-box access to query the model $f_{tar}$ without knowing the details of the target model $f_{tar}$ architecture or parameters like gradient information $\nabla_{Tr}$. The attacker attempts to compute  training data reconstruction (i.e., ImRec) attack without having access to other additional information, e.g., blurred or masked images $D_{s}$.

Two major components of the model inversion attack workflow are the target model $f_{tar}$ and the surrogate attack model $f_{sar}$~\cite{jia2018attriguard,dibbo2023sok,zhao2021feasibility}. 
Training data reconstruction (i.e., ImRec) attack in the literature considers the target model $f_{tar}$ to be either the split network~\cite{titcombe2021practical} or the end-to-end network~\cite{gong2023gan,zhang2020secret}. In the split network $f_{tar}$ model, the output of a particular layer $l$ in the network, i.e., $a^{[l]}$, where $1\leq l < L$ is accessible to the adversary, whereas, for the end-to-end network, the adversary does not have access to intermediate layer outputs; rather, the adversary only has access to the output from the last hidden layer before the classification layer $a^{[L]}$.

\section{Results of extra \{threat model, dataset\} experiments}
\label{app:otherexperiments}
We experiment all 3 attack setups: \emph{Plug-\&-Play} model inversion attack~\cite{struppek2022plug}, \emph{end-to-end}, and \emph{split} on three additional datasets: MNIST, Fashion MNIST, and CIFAR10. We experiment with all benchmarks and present the results on Tables~\ref{table:performance_p_n_p_attack}, ~\ref{table:performanceetn},  and~\ref{table:performancesplit}. In all of these additional datasets, \textsc{SCA} consistently outperforms all benchmarks.

\begin{table}[h!]
\centering
\caption{Experiments set $1$ \textbf{Additional Datasets}: Performance in Plug-\&-Play Model Inversion Attack~\cite{struppek2022plug} setting \emph{(lower rows=better defense)}.} 
\begin{tabular}{ p{1.3cm} p{3.5cm} p{1.8cm} p{1.5cm} p{2.0cm} p{1.5cm}  }

\hline
 Dataset & Defense   & PSNR $\downdownarrows$    &SSIM $\downdownarrows$  &\mbox{FID $\upuparrows$}  & Accuracy\\
\hline

\hline
MNIST& \textsc{No-Defense}  &$7.24	$&$0.783	$&$23.6	$&$0.971$\\ 
 & \textsc{Gaussian-Noise}  &$6.94$	&$0.686	$&$31.22	$&$0.958 $\\
 & \textsc{GAN}  &$6.83$	&$0.734$	&$89.38	$&$0.968 $\\
     & \mbox{Gong et al.\cite{gong2023gan}}++  &$6.69$	&$0.716$	&$92.21$	&$0.987$ \\
  & \mbox{Titcombe et al.\cite{titcombe2021practical}}  &$6.34$	&$0.744$	&$131.8$	&$0.980$ \\
    &  \mbox{Gong et al.\cite{gong2023gan}}  &$6.76$	&$0.681$	&$99.53$	&$0.985$ \\
   & \mbox{Peng et al.\cite{peng2022bilateral}}  &$6.89 $&$0.704  $&$283.8  $&$0.941$\\
    & \mbox{Hayes et al.\cite{hayes2023bounding}}  &$7.03	$&$0.672	$&$396.1	$&$0.871$\\
 
    & \mbox{Wang et al.\cite{wang2021improving}}  &$7.14	$&$0.752	$&$261.2	$&$0.937$\\
   & \textsc{Sparse-Standard}   &$6.24$	&$0.631$	&$158.6$		&$0.986$ \\
& \textbf{\textsc{SCA0.1}} & $\mathbf{6.19}$	&$\mathbf{0.633}$	&$\mathbf{287.9}$ & $\mathbf{0.984}$ \\ 
& \textbf{\mbox{\textsc{SCA0.25}}} & $\mathbf{5.83}$	&$\mathbf{0.607}$	&$\mathbf{289.3}$ & $\mathbf{0.983}$ \\ 
& \textbf{\textsc{SCA0.5}} & $\mathbf{5.74}$	&$\mathbf{0.604}$	&$\mathbf{299.6}$ & $\mathbf{0.977}$ \\ 
\hline

\hline
\hline
Fashion & \textsc{No-Defense}  &$8.91	$&$0.147	$&$235.5	$&$0.886$\\ 
 MNIST& \textsc{Gaussian-Noise}  &$8.67$	&$0.132	$&$239.8	$&$0.815 $\\
 & \textsc{GAN}  &$8.66$	&$0.147$	&$243.3	$&$0.883 $\\
     & \mbox{Gong et al.\cite{gong2023gan}}++  &$8.73$	&$0.130$	&$220.2$	&$0.906$ \\
  & \mbox{Titcombe et al.\cite{titcombe2021practical}}  &$8.56$	&$0.134$	&$229.8$	&$0.905$ \\
    &  \mbox{Gong et al.\cite{gong2023gan}}  &$8.57$	&$0.143$	&$244.3$	&$0.888$ \\
    & \mbox{Peng et al.\cite{peng2022bilateral}}  &$8.85 $&$0.147  $&$227.5  $&$0.845$\\
& \mbox{Hayes et al.\cite{hayes2023bounding}}  &$8.63	$&$0.139	$&$218.4	$&$0.752$\\
    
     & \mbox{Wang et al.\cite{wang2021improving}}  &$8.90	$&$0.119	$&$210.3	$&$0.880$\\
   & \textsc{Sparse-Standard}   &$8.71$	&$0.135$	&$223.3$		&$0.879$ \\
& \textbf{\textsc{SCA0.1}} & $\mathbf{8.49}$	&$\mathbf{0.039}$	&$\mathbf{222.8}$ & $\mathbf{0.897}$ \\ 
& \textbf{\mbox{\textsc{SCA0.25}}} & $\mathbf{8.49}$	&$\mathbf{0.032}$	&$\mathbf{229.9}$ & $\mathbf{0.887}$ \\ 
& \textbf{\textsc{SCA0.5}} & $\mathbf{8.45}$	&$\mathbf{0.047}$	&$\mathbf{233.5}$ & $\mathbf{0.876}$ \\ 
\hline
\hline

CIFAR10&  \textsc{No-Defense}  &$11.94$	&$0.381	$&$39.38	$&$0.821$\\ 
& \textsc{Gaussian-Noise}  &$11.88	$&$0.365	$&$77.92	$&$0.626$ \\
 & \textsc{GAN}  &$11.86$	&$0.369	$&$88.39	$&$0.596$ \\
  & \mbox{Titcombe et al.\cite{titcombe2021practical}}   &$10.89	$&$0.346	$&$79.19	$&$0.792$\\

  & \mbox{Gong et al.\cite{gong2023gan}}++  &$11.06	$&$0.339	$&$78.48	$&$0.773$\\
  & \mbox{Gong et al.\cite{gong2023gan}}  &$11.21	$&$0.334	$&$92.33	$&$0.682$\\
    & \mbox{Peng et al.\cite{peng2022bilateral}}  &$11.96 $&$0.354  $&$120.5  $&$0.752$\\
    & \mbox{Hayes et al.\cite{hayes2023bounding}}  &$11.12	$&$0.342	$&$142.1	$&$0.626$\\
 
    & \mbox{Wang et al.\cite{wang2021improving}}  &$11.02	$&$0.346	$&$142.6	$&$0.756$\\
    & \textsc{Sparse-Standard}  &$10.74	$&$0.303	$&$137.4	$&$0.790$ \\
& \textbf{\textsc{SCA0.1}} & $\mathbf{10.59}$	&$\mathbf{0.305}$	&$\mathbf{144.1}$ & $\mathbf{0.787}$ \\ 
& \textbf{\mbox{\textsc{SCA0.25}}} & $\mathbf{10.27}$	&$\mathbf{0.279}$	&$\mathbf{189.9}$ & $\mathbf{0.772}$ \\
& \textbf{\textsc{SCA0.5}} & $\mathbf{10.23}$	&$\mathbf{0.276}$	&$\mathbf{189.7}$ & $\mathbf{	0.744}$ \\ 

\hline

\hline

\end{tabular}

\label{table:performance_p_n_p_attack}
\end{table}

\begin{table}[h!]
\centering
\caption{Experiments set $2$  \textbf{Additional Datasets}: Performance in \emph{end-to-end} network setting \emph{(lower rows=better defense)}.}
\begin{tabular}{ p{1.3cm} p{3.5cm} p{1.8cm} p{1.5cm} p{2.0cm} p{1.5cm}  }
\hline
 Dataset & Defense   & PSNR $\downdownarrows$    &SSIM $\downdownarrows$  &\mbox{FID  $\upuparrows$}  & Accuracy\\
\hline
MNIST&  \textsc{No-Defense}  &$40.87$	&$0.982	$&$16.31	$&$0.971$\\ 
& \textsc{Gaussian-Noise}  &$40.88	$&$0.983	$&$15.88	$&$0.958$ \\
 & \textsc{GAN}  &$40.69$	&$0.981	$&$16.59	$&$0.968$ \\
  & \mbox{Titcombe et al.\cite{titcombe2021practical}}   &$31.18	$&$0.863	$&$47.32	$&$0.980$\\

  & \mbox{Gong et al.\cite{gong2023gan}}++  &$30.37	$&$0.838	$&$72.99	$&$0.987$\\
  & \mbox{Gong et al.\cite{gong2023gan}}  &$29.05	$&$0.817	$&$75.39	$&$0.985$\\
    & \mbox{Peng et al.\cite{peng2022bilateral}}  &$18.44 $&$0.354  $&$111.6  $&$0.968$\\
& \mbox{Hayes et al.\cite{hayes2023bounding}}  &$19.75	$&$0.488	$&$298.8	$&$0.871$\\
        & \mbox{Wang et al.\cite{wang2021improving}}  &$27.26	$&$0.862	$&$72.66	$&$0.962$\\
    & \textsc{Sparse-Standard}  &$21.34	$&$0.439	$&$142.9	$&$0.986$ \\
& \textbf{\textsc{SCA0.1}} & $\mathbf{19.54}$	&$\mathbf{0.502}$	&$\mathbf{178.5}$ & $\mathbf{0.984}$ \\ 
& \textbf{\mbox{\textsc{SCA0.25}}} & $\mathbf{18.81}$	&$\mathbf{0.340}$	&$\mathbf{174.1}$ & $\mathbf{0.983}$ \\ 
& \textbf{\textsc{SCA0.5}} & $\mathbf{17.85}$	&$\mathbf{0.164}$	&$\mathbf{335.5}$ & $\mathbf{0.977}$ \\ 

\hline
\hline
Fashion& \textsc{No-Defense}  &$37.86	$&$0.975	$&$13.91	$&$0.886 $\\ 
MNIST & \textsc{Gaussian-Noise}  &$36.54$	&$0.969	$&$16.49	$&$0.815 $\\
 & \textsc{GAN}  &$37.68$	&$0.974$	&$19.26	$&$0.883 $\\
     & \mbox{Gong et al.\cite{gong2023gan}}++  &$27.71$	&$0.794$	&$41.35$	&$0.906$ \\
  & \mbox{Titcombe et al.\cite{titcombe2021practical}}  &$26.66$	&$0.759$	&$53.76$	&$0.905$ \\
    &  \mbox{Gong et al.\cite{gong2023gan}}  &$21.24$	&$0.523$	&$93.08$	&$0.888$ \\
   & \mbox{Peng et al.\cite{peng2022bilateral}}  &$17.98 $&$0.368  $&$70.53  $&$0.880$\\
& \mbox{Hayes et al.\cite{hayes2023bounding}}  &$21.13	$&$0.297	$&$223.3	$&$0.752$\\
   
       & \mbox{Wang et al.\cite{wang2021improving}}  &$25.98	$&$0.806	$&$41.87	$&$0.838$\\
   & \textsc{Sparse-Standard}   &$19.35$	&$0.446$	&$128.4$		&$0.879$ \\
& \textbf{\textsc{SCA0.1}} & $\mathbf{17.92}$	&$\mathbf{0.209}$	&$\mathbf{196.1}$ & $\mathbf{0.897}$ \\ 
& \textbf{\mbox{\textsc{SCA0.25}}} & $\mathbf{17.03}$	&$\mathbf{0.186}$	&$\mathbf{195.2}$ & $\mathbf{0.887}$ \\ 
& \textbf{\textsc{SCA0.5}} & $\mathbf{14.51}$	&$\mathbf{0.069}$	&$\mathbf{423.2}$ & $\mathbf{0.876}$ \\ 
\hline
\hline
CIFAR10&  \textsc{No-Defense}  &$21.17$	&$0.477	$&$70.96	$&$0.821$\\ 
& \textsc{Gaussian-Noise}  &$20.26	$&$0.220	$&$77.42	$&$0.626$ \\
 & \textsc{GAN}  &$19.71$	&$0.259	$&$132.0	$&$0.596$ \\
  & \mbox{Titcombe et al.\cite{titcombe2021practical}}   &$18.62	$&$0.174	$&$171.9	$&$0.792$\\

  & \mbox{Gong et al.\cite{gong2023gan}}++  &$18.27	$&$0.209	$&$149.1	$&$0.773$\\
  & \mbox{Gong et al.\cite{gong2023gan}}  &$19.10	$&$0.150	$&$133.8	$&$0.682$\\
    & \mbox{Peng et al.\cite{peng2022bilateral}}  &$17.20 $&$0.002  $&$130.3  $&$0.717$\\

    & \mbox{Hayes et al.\cite{hayes2023bounding}}  &$17.95	$&$0.002	$&$142.4	$&$0.626$\\
    
    & \mbox{Wang et al.\cite{wang2021improving}}  &$17.08	$&$0.002	$&$136.1	$&$0.793$\\
    & \textsc{Sparse-Standard}  &$18.01	$&$0.003	$&$168.6	$&$0.790$ \\
& \textbf{\textsc{SCA0.1}} & $\mathbf{17.09}$	&$\mathbf{0.001}$	&$\mathbf{172.0}$ & $\mathbf{0.787}$ \\ 
& \textbf{\mbox{\textsc{SCA0.25}}} & $\mathbf{16.78}$	&$\mathbf{0.001}$	&$\mathbf{189.3}$ & $\mathbf{0.772}$ \\
& \textbf{\textsc{SCA0.5}} & $\mathbf{16.24}$	&$\mathbf{0.001}$	&$\mathbf{197.0}$ & $\mathbf{0.744}$ \\ 
\hline
\end{tabular}
\label{table:performanceetn}
\end{table}

\begin{table}[h!]
\centering
\caption{Experiments set $3$  \textbf{Additional Datasets}: Performance in \emph{split} network setting \emph{(lower rows=better defense)}.}
\begin{tabular}{ p{1.3cm} p{3.5cm} p{1.8cm} p{1.5cm} p{2.0cm} p{1.5cm}  }

\hline
 Dataset & Defense   & PSNR $\downdownarrows$    &SSIM $\downdownarrows$  &\mbox{FID  $\upuparrows$}  &Accuracy \\
\hline
MNIST&  \textsc{No-Defense}  &$31.21$	&$0.923$	&$19.64$	&$0.963$\\ 
 & \textsc{Gaussian-Noise}  &$31.07$	&$0.922$	&$23.27$	&$0.972$ \\

 & \textsc{GAN}  &$28.39$	&$0.894	$&$27.26$	&$0.969$ \\
 & \mbox{Gong et al.\cite{gong2023gan}}  &$28.30	$&$0.806	$&$69.38	$&$0.986$\\
 & \mbox{Titcombe et al.\cite{titcombe2021practical}}  &$25.40	$&$0.713$	&$76.88$	&$0.952$ \\

  & \mbox{Gong et al.\cite{gong2023gan}}++  &$21.94$	&$0.591	$&$97.33	$&$0.991$\\

  & \mbox{Peng et al.\cite{peng2022bilateral}}  &$16.90 $&$0.475  $&$103.2  $&$0.960$\\

 & \mbox{Hayes et al.\cite{hayes2023bounding}}  &$17.23	$&$0.030	$&$288.1	$&$0.856$\\
   
  & \mbox{Wang et al.\cite{wang2021improving}}  &$21.87	$&$0.696	$&$53.09	$&$0.903$\\
 & \textsc{Sparse-Standard}&$18.71	$&$0.288$	&$188.4$	&$0.981$ \\
& \textbf{\textsc{SCA0.1}} & $\mathbf{16.17}$	&$\mathbf{0.109}$	&$\mathbf{227.4}$ & $\mathbf{0.988}$ \\
& \textbf{\mbox{\textsc{SCA0.25}}} & $\mathbf{17.40}$	&$\mathbf{0.058}$	&$\mathbf{301.6}$ & $\mathbf{0.980}$ \\
& \textbf{\textsc{SCA0.5}} & $\mathbf{14.98}$	&$\mathbf{0.044}$	&$\mathbf{307.7}$ & $\mathbf{0.975}$ \\ 

\hline
\hline
\mbox{Fashion}&  \textsc{No-Defense}  &$29.66$	&$0.911	$&$14.33	$&$0.868$ \\ 
 MNIST & \textsc{Gaussian-Noise}  &$29.49$	&$0.909	$&$14.81	$&$0.871$ \\
 & \textsc{GAN}  &$26.03$	&$0.849	$&$19.33	$&$0.885$ \\
   
    &  \mbox{Gong et al.\cite{gong2023gan}}  &$23.70$	&$0.631$	&$97.52$	&$0.884$ \\
  & \mbox{Titcombe et al.\cite{titcombe2021practical}}  &$20.48$	&$0.565	$&$81.01$	&$0.872$ \\
   & \mbox{Gong et al.\cite{gong2023gan}}++  &$25.77$	&$0.726$	&$57.72$	&$0.908$\\

 & \mbox{Peng et al.\cite{peng2022bilateral}}  &$20.67 $&$0.583  $&$46.48  $&$0.865$\\

& \mbox{Hayes et al.\cite{hayes2023bounding}}  &$20.10	$&$0.256	$&$200.6	$&$0.748$\\
  
 & \mbox{Wang et al.\cite{wang2021improving}}  &$24.53	$&$0.588	$&$81.79	$&$0.881$\\

     & \textsc{Sparse-Standard}  &$19.54$	&$0.405	$&$200.5	$&$0.882$ \\
& \textbf{\textsc{SCA0.1}} & $\mathbf{18.11}$	&$\mathbf{0.154}$	&$\mathbf{171.1}$ & $\mathbf{0.904}$ \\
& \textbf{\mbox{\textsc{SCA0.25}}} & $\mathbf{17.74}$	&$\mathbf{0.188}$	&$\mathbf{203.8}$ & $\mathbf{0.896}$ \\
& \textbf{\textsc{SCA0.5}} & $\mathbf{17.15}$	&$\mathbf{0.134}$	&$\mathbf{270.4}$ & $\mathbf{0.879}$ \\ 
\hline
\hline
CIFAR10&  \textsc{No-Defense}  &$16.48$	&$0.709	$&$47.77	$&$0.823$\\ 
& \textsc{Gaussian-Noise}  &$14.79	$&$0.311	$&$149.5	$&$0.598$ \\
 & \textsc{GAN}  &$14.87$	&$0.296	$&$13.01	$&$0.675$ \\
  & \mbox{Titcombe et al.\cite{titcombe2021practical}}   &$14.68	$&$0.244	$&$157.3	$&$0.779$\\

  & \mbox{Gong et al.\cite{gong2023gan}}++  &$13.32	$&$0.003	$&$162.4	$&$0.691$\\
  & \mbox{Gong et al.\cite{gong2023gan}}  &$14.55	$&$0.291	$&$152.1	$&$0.644$\\
     & \mbox{Peng et al.\cite{peng2022bilateral}}  &$17.18 $&$0.002  $&$169.1  $&$0.707$\\
    & \mbox{Hayes et al.\cite{hayes2023bounding}}  &$15.44	$&$0.005	$&$204.5	$&$0.596$\\
    & \mbox{Wang et al.\cite{wang2021improving}}  &$14.73	$&$0.001	$&$176.3	$&$0.820$\\
    & \textsc{Sparse-Standard}  &$13.22	$&$0.003	$&$167.9	$&$0.769$ \\
& \textbf{\textsc{SCA0.1}} & $\mathbf{13.18}$	&$\mathbf{0.002}$	&$\mathbf{174.2}$ & $\mathbf{0.758}$ \\ 
& \textbf{\mbox{\textsc{SCA0.25}}} & $\mathbf{13.07}$	&$\mathbf{0.002}$	&$\mathbf{181.2}$ & $\mathbf{0.742}$ \\ 
& \textbf{\textsc{SCA0.5}} & $\mathbf{12.88}$	&$\mathbf{0.002}$	&$\mathbf{375.3}$ & $\mathbf{	0.739}$ \\ 

\hline
\end{tabular}
\label{table:performancesplit}
\end{table}

\begin{table}[h!]
\centering
\caption{Experiments set $1$: additional Laplace noise benchmark with larger $1.0$ noise parameter: Performance in Plug-\&-Play Model Inversion Attack~\cite{struppek2022plug} setting \emph{(lower rows=better defense)}.}
\begin{tabular}{ p{1.3cm} p{3.5cm} p{1.8cm} p{1.5cm} p{2.0cm} p{1.5cm}  }
\hline
 Dataset & Defense   & PSNR $\downdownarrows$    &SSIM $\downdownarrows$  &\mbox{FID  $\upuparrows$}  & Accuracy\\
\hline
MNIST&   \mbox{Titcombe et al.\cite{titcombe2021practical}-$1.0$}  &$6.60	$&$0.685	$&$280.1	$&$0.938$\\

   & \textbf{\textsc{SCA0.1}} & $\mathbf{6.19}$	&$\mathbf{0.633}$	&$\mathbf{287.9}$ & $\mathbf{0.984}$ \\ 
& \textbf{\mbox{\textsc{SCA0.25}}} & $\mathbf{5.83}$	&$\mathbf{0.607}$	&$\mathbf{289.3}$ & $\mathbf{0.983}$ \\ 
& \textbf{\textsc{SCA0.5}} & $\mathbf{5.74}$	&$\mathbf{0.604}$	&$\mathbf{299.6}$ & $\mathbf{0.977}$ \\ 

\hline
\hline
Fashion &    \mbox{Titcombe et al.\cite{titcombe2021practical}-$1.0$}  &$8.72$	&$0.1412$	&$232.1$	&$0.823$ \\

MNIST&  \textbf{\textsc{SCA0.1}} & $\mathbf{8.49}$	&$\mathbf{0.039}$	&$\mathbf{222.8}$ & $\mathbf{0.897}$ \\ 
& \textbf{\mbox{\textsc{SCA0.25}}} & $\mathbf{8.49}$	&$\mathbf{0.032}$	&$\mathbf{229.9}$ & $\mathbf{0.887}$ \\ 
& \textbf{\textsc{SCA0.5}} & $\mathbf{8.45}$	&$\mathbf{0.047}$	&$\mathbf{233.5}$ & $\mathbf{0.876}$ \\ 
\hline
\hline
CIAFR10 &    \mbox{Titcombe et al.\cite{titcombe2021practical}-$1.0$}  &$10.75$	&$0.335$	&$112.7$	&$0.779$ \\

& \textbf{\textsc{SCA0.1}} & $\mathbf{10.59}$	&$\mathbf{0.305}$	&$\mathbf{144.1}$ & $\mathbf{0.787}$ \\ 
& \textbf{\mbox{\textsc{SCA0.25}}} & $\mathbf{10.27}$	&$\mathbf{0.279}$	&$\mathbf{189.9}$ & $\mathbf{0.772}$ \\
& \textbf{\textsc{SCA0.5}} & $\mathbf{10.23}$	&$\mathbf{0.276}$	&$\mathbf{189.7}$ & $\mathbf{	0.744}$ \\ 
\hline
\end{tabular}
\label{table:extra_three_pnp}
\end{table}

\begin{table}[h!]
\centering
\caption{Experiments set $2$ additional Laplace noise benchmark with larger $1.0$ noise parameter: Performance in \emph{end-to-end} network setting \emph{(lower rows=better defense)}.}
\begin{tabular}{ p{1.3cm} p{3.5cm} p{1.8cm} p{1.5cm} p{2.0cm} p{1.5cm}  }

\hline
 Dataset & Defense   & PSNR $\downdownarrows$    &SSIM $\downdownarrows$  &\mbox{FID  $\upuparrows$}  & Accuracy\\
\hline
MNIST&  \mbox{Titcombe et al.\cite{titcombe2021practical}-$1.0$}  &$24.89	$&$0.664	$&$50.64	$&$0.938$\\
& \textbf{\textsc{SCA0.1}} & $\mathbf{19.54}$	&$\mathbf{0.502}$	&$\mathbf{178.5}$ & $\mathbf{0.984}$ \\ 
   
& \textbf{\mbox{\textsc{SCA0.25}}} & $\mathbf{18.81}$	&$\mathbf{0.340}$	&$\mathbf{174.1}$ & $\mathbf{0.983}$ \\ 
& \textbf{\textsc{SCA0.5}} & $\mathbf{17.85}$	&$\mathbf{0.164}$	&$\mathbf{335.5}$ & $\mathbf{0.977}$ \\ 

\hline
\hline
Fashion&    \mbox{Titcombe et al.\cite{titcombe2021practical}-$1.0$}  &$20.21$	&$0.567$	&$80.55$	&$0.823$ \\
MNIST& \textbf{\textsc{SCA0.1}} & $\mathbf{17.92}$	&$\mathbf{0.209}$	&$\mathbf{196.1}$ & $\mathbf{0.897}$ \\

& \textbf{\mbox{\textsc{SCA0.25}}} & $\mathbf{17.03}$	&$\mathbf{0.186}$	&$\mathbf{195.2}$ & $\mathbf{0.887}$ \\ 
& \textbf{\textsc{SCA0.5}} & $\mathbf{14.51}$	&$\mathbf{0.069}$	&$\mathbf{423.2}$ & $\mathbf{0.876}$ \\ 
\hline
\hline
CIFAR10&    \mbox{Titcombe et al.\cite{titcombe2021practical}-$1.0$}  &$18.71$	&$0.672$	&$170.8$	&$0.779$ \\
& \textbf{\textsc{SCA0.1}} & $\mathbf{17.09}$	&$\mathbf{0.001}$	&$\mathbf{172.0}$ & $\mathbf{0.787}$ \\ 
& \textbf{\mbox{\textsc{SCA0.25}}} & $\mathbf{16.78}$	&$\mathbf{0.001}$	&$\mathbf{189.3}$ & $\mathbf{0.772}$ \\
& \textbf{\textsc{SCA0.5}} & $\mathbf{16.24}$	&$\mathbf{0.001}$	&$\mathbf{197.0}$ & $\mathbf{0.744}$ \\ 
\hline
\end{tabular}
\label{table:extra_one}
\end{table}

\section{Additional baseline tuning}
\label{appendix:addlexp}
\begin{figure*}[]
\centering
  \centering
  \includegraphics[width=12cm]
  {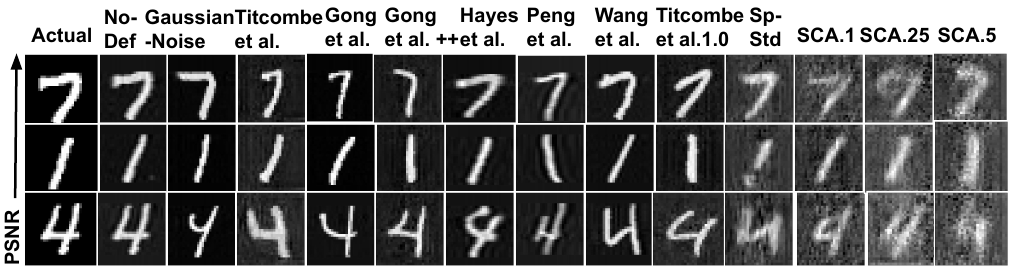}
\caption{Qualitative comparisons among actual and reconstructed images under \textsc{SCA} and additional Laplace noise defense benchmark with larger $1.0$ noise parameter. 
}
\label{fig:recon_comparison2}
\end{figure*}

We also attempt to improve the Laplace noise-based defense of Titcombe et al. \cite{titcombe2021practical} by increasing the noise scale parameter $b$ from $\mathcal{L}(\mu$$=$$0$, $b$$=$$0.5)$ to $\mathcal{L}(\mu$$=$$0$, $b$$=$$1.0)$. Tables \ref{table:extra_three_pnp}, \ref{table:extra_one}, and \ref{table:extra_two} compare these results to \textsc{SCA} for in all 3 attack settings. Observe that the additional noise significantly degrades classification accuracy in all but one case, yet it does not result in reconstruction metrics that rival those of \textsc{SCA}'s. In Figure~\ref{fig:recon_comparison2}, we present the reconstructed images in the Split network attack setting on MNIST data. We also include the Laplace noise-based defense with higher noise parameter $\mathcal{L}(\mu$$=$$0$, $b$$=$$1.0)$.

\begin{table}[h!]
\centering
\caption{Experiments set $3$: additional Laplace noise benchmark with larger $1.0$ noise parameter: Performance in \emph{split} network setting \emph{(lower rows=better defense)}.}
\begin{tabular}{ p{1.3cm} p{3.5cm} p{1.8cm} p{1.5cm} p{2.0cm} p{1.5cm}  }
\hline
 Dataset & Defense   & PSNR $\downdownarrows$    &SSIM $\downdownarrows$  &\mbox{FID  $\upuparrows$}  & Accuracy\\
\hline
MNIST&   \mbox{Titcombe et al.\cite{titcombe2021practical}-$1.0$}  &$22.63	$&$0.503	$&$66.40	$&$0.980$\\

   & \textbf{\textsc{SCA0.1}} & $\mathbf{16.17}$	&$\mathbf{0.109}$	&$\mathbf{227.4}$ & $\mathbf{0.988}$ \\
& \textbf{\mbox{\textsc{SCA0.25}}} & $\mathbf{17.40}$	&$\mathbf{0.058}$	&$\mathbf{301.6}$ & $\mathbf{0.980}$ \\
& \textbf{\textsc{SCA0.5}} & $\mathbf{14.98}$	&$\mathbf{0.044}$	&$\mathbf{307.7}$ & $\mathbf{0.975}$ \\ 

\hline
\hline
Fashion &    \mbox{Titcombe et al.\cite{titcombe2021practical}-$1.0$}  &$18.36$	&$0.408$	&$80.80$	&$0.878$ \\

MNIST& \textbf{\textsc{SCA0.1}} & $\mathbf{18.11}$	&$\mathbf{0.154}$	&$\mathbf{171.1}$ & $\mathbf{0.904}$ \\
 & \textbf{\mbox{\textsc{SCA0.25}}} & $\mathbf{17.74}$	&$\mathbf{0.188}$	&$\mathbf{203.8}$ & $\mathbf{0.896}$ \\
& \textbf{\textsc{SCA0.5}} & $\mathbf{17.15}$	&$\mathbf{0.134}$	&$\mathbf{270.4}$ & $\mathbf{0.879}$ \\ 
\hline
\hline
CIAFR10 &    \mbox{Titcombe et al.\cite{titcombe2021practical}-$1.0$}  &$14.27$	&$0.259$	&$171.6$	&$0.786$ \\

& \textbf{\textsc{SCA0.1}} & $\mathbf{13.18}$	&$\mathbf{0.002}$	&$\mathbf{174.2}$ & $\mathbf{0.758}$ \\ 
& \textbf{\mbox{\textsc{SCA0.25}}} & $\mathbf{13.07}$	&$\mathbf{0.002}$	&$\mathbf{181.2}$ & $\mathbf{0.742}$ \\ 
& \textbf{\textsc{SCA0.5}} & $\mathbf{12.88}$	&$\mathbf{0.002}$	&$\mathbf{375.3}$ & $\mathbf{	0.739}$ \\ 
\hline
\end{tabular}
\label{table:extra_two}
\end{table}

\section{Stability analysis of \textsc{SCA}}
\label{appendix:stability_sca}
 Tables~\ref{table:all_performance_celeba_pap} and ~\ref{table:all_performance_medmnist_etn} 
 show mean metrics and std. deviation error bars taken over \emph{multiple runs} of each defense. Observe that SCA is at least as stable (and in some cases significantly more stable) than alternatives.

\begin{table}[h!]
\centering
\caption{\textbf{Stability analysis 1:} Performance comparison (mean$\pm$ standard deviations) across multiple runs in Plug-\&-Play Model Inversion Attack~\cite{struppek2022plug} setting \emph{(lower rows=better defense)} on high-res CelebA dataset.}
\begin{tabular}{ p{1.3cm} p{2.8cm} p{1.8cm} p{1.8cm} p{1.8cm} p{1.8cm}  }

\hline
 Dataset & Defense   & PSNR $\downdownarrows$    &SSIM $\downdownarrows$  &\mbox{FID  $\upuparrows$}  & Accuracy\\
\hline

   CelebA& \textsc{No-Defense}  &$11.42\pm 2.44	$&$0.613\pm0.29	$&$292.9\pm 81.5	$&$0.721\pm 0.04$\\ 
 & \textsc{Gaussian-Noise}  &$10.87\pm 2.25	$&$0.614\pm 0.30	$&$296.5\pm 73.3	$&$0.624\pm 0.03$\\
 & \textsc{GAN}  &$11.02\pm 1.82	$&$0.600\pm 0.29	$&$301.4\pm 92.4	$&$0.613\pm 0.02$\\ 
     & \mbox{Gong et al.\cite{gong2023gan}}++  &$10.84\pm 1.94	$&$0.556\pm 0.28	$&$301.0\pm 81.4	$&$0.658\pm 0.02$\\ 
  & \mbox{Titcombe et al.\cite{titcombe2021practical}}  &$10.76\pm 2.37	$&$0.557\pm 0.24	$&$345.5\pm 86.1	$&$0.643\pm 0.01$\\ 
    &  \mbox{Gong et al.\cite{gong2023gan}}  &$10.91\pm 1.88	$&$0.560\pm 0.29	$&$304.5\pm 82.5	$&$0.616\pm 0.01$\\ 
   & \mbox{Peng et al.\cite{peng2022bilateral}}  &$10.17\pm 2.32	$&$0.491\pm 0.24	$&$399.1\pm 55.3	$&$0.667\pm 0.04$\\ 
& \mbox{Hayes et al.\cite{hayes2023bounding}}  &$10.16\pm 1.95	$&$0.535\pm 0.25	$&$320.8\pm 79.0	$&$0.601\pm 0.02$\\ 
    & \mbox{Wang et al.\cite{wang2021improving}}  &$10.39\pm 2.55	$&$0.505\pm 0.24	$&$341.7\pm 74.2	$&$0.669\pm 0.05$\\ 
   & \textsc{Sparse-Std}   &$9.78\pm 2.13	$&$0.485\pm 0.24	$&$367.3\pm 44.7	$&$0.663\pm 0.03$\\ 
& \textbf{\textsc{SCA0.1}} &$\mathbf{9.56} \pm 2.30$	&$\mathbf{0.454}\pm 0.25$	&$\mathbf{396.6}\pm 45.0$ & $\mathbf{0.659}\pm 0.04$ \\
& \textbf{\mbox{\textsc{SCA0.25}}} &$\mathbf{9.27} \pm 2.06$	&$\mathbf{0.452}\pm 0.25$	&$\mathbf{412.8}\pm 53.7$ & $\mathbf{0.661}\pm 0.05$ \\
& \textbf{\textsc{SCA0.5}} &$\mathbf{9.12} \pm 2.68$	&$\mathbf{0.368}\pm 0.24$	&$\mathbf{421.7}\pm 49.9$ & $\mathbf{0.653}\pm 0.04$ \\ 
\hline
\end{tabular}
\label{table:all_performance_celeba_pap}
\end{table}

\begin{table}[h!]
\centering
\caption{\textbf{Stability analysis 2:} Performance comparison (mean$\pm$ standard deviations) across multiple runs in \emph{end-to-end} network setting \emph{(lower rows=better defense)} on Medical MNIST dataset.}
\begin{tabular}{ p{1.3cm} p{2.8cm} p{1.8cm} p{1.8cm} p{1.8cm} p{1.8cm}  }

\hline
 Dataset & Defense   & PSNR $\downdownarrows$    &SSIM $\downdownarrows$  &\mbox{FID  $\upuparrows$}  & Accuracy\\
\hline

Medical & \textsc{No-Defense}  &$30.17\pm 0.90	$&$0.912\pm 0.01	$&$12.40\pm 8.69	$&$0.998\pm 0.01$\\ 
MNIST & \textsc{Gaussian-Noise}  &$27.00\pm 1.30	$&$0.828\pm 0.05	$&$17.29\pm 11.9	$&$0.886\pm 0.06$\\ 
 & \textsc{GAN}  &$25.05\pm 2.78	$&$0.699\pm 0.03	$&$29.08\pm 20.5	$&$0.995\pm 0.01$\\ 
     & \mbox{Gong et al.\cite{gong2023gan}}++  &$20.37\pm 1.65	$&$0.451\pm 0.03	$&$44.68\pm 30.9	$&$0.871\pm 0.01$\\ 
  & \mbox{Titcombe et al.\cite{titcombe2021practical}}  &$20.51\pm 0.28	$&$0.574\pm 0.01	$&$28.23\pm 1.65	$&$0.805\pm 0.06$\\ 
    &  \mbox{Gong et al.\cite{gong2023gan}}  &$23.65\pm 1.07	$&$0.605\pm 0.09	$&$37.16\pm 26.2	$&$0.757\pm 0.03$\\ 
   & \mbox{Peng et al.\cite{peng2022bilateral}}  &$17.42\pm 2.87	$&$0.519\pm 0.22	$&$65.39\pm 32.8	$&$0.866\pm 0.08$\\ 
& \mbox{Hayes et al.\cite{hayes2023bounding}}  &$19.57\pm 1.08	$&$0.003\pm 0.01	$&$155.0\pm 92.5	$&$0.847\pm 0.08$\\ 
    & \mbox{Wang et al.\cite{wang2021improving}}  &$17.89\pm 2.09	$&$0.463\pm 0.08	$&$101.8\pm 66.5	$&$0.829\pm 0.08$\\ 
   & \textsc{Sparse-Std}   &$13.49\pm 0.29	$&$0.158\pm 0.09	$&$203.4\pm 92.2	$&$0.865\pm 0.05$\\ 
& \textbf{\textsc{SCA0.1}} &$\mathbf{12.46} \pm 0.30$	&$\mathbf{0.006}\pm 0.01$	&$\mathbf{231.8}\pm 124$ & $\mathbf{0.858}\pm 0.08$ \\
& \textbf{\mbox{\textsc{SCA0.25}}} &$\mathbf{11.89} \pm 0.35$	&$\mathbf{0.008}\pm 0.01$	&$\mathbf{254.1}\pm 153$ & $\mathbf{0.850}\pm 0.08$ \\
& \textbf{\textsc{SCA0.5}} &$\mathbf{11.19} \pm 0.11$	&$\mathbf{0.001}\pm 0.01$	&$\mathbf{276.9}\pm 97.0$ & $\mathbf{0.841}\pm 0.08$ \\ 
\hline
\end{tabular}
\label{table:all_performance_medmnist_etn}
\end{table}

\section{Compute time}
\label{app:time}
Our basic \textsc{SCA} research implementation completes in comparable or less compute time than highly optimized implementations of benchmarks. In the `worst-case' across all of our experiments, \textsc{SCA} is faster than the best performing baseline (Peng et al. \cite{peng2022bilateral}) but slower than other baselines. Table \ref{tab:computetimes} shows the compute times (in seconds) for this `worst-case' experiment below (The MNIST dataset under the Plug-\&-Play attack~\cite{struppek2022plug}).

\section{Ablations: Tuning \textsc{SCA}}
\label{tuning:sca}
Observe that our \textsc{SCA} outperforms SOTA defense baselines in robustness even without any tuning of parameters. However, tuning the hyper-parameters can boost the accuracy further, e.g., we use kernel size as default 5 for all experiments. Increasing the kernel from 5 to 7 can improve \textsc{SCA} accuracies beyond. While our goal is to capture the essence of the \textsc{SCA} itself in terms of robustness, we explore a little bit on further possible improvements on accuracy scores. We consider the lowest robust \textsc{SCA}, i.e., \textsc{SCA0.1} for the tuning of kernel size, and we present the comparisons of accuracies between \textsc{SCA0.1} and \textsc{Tuned SCA0.1} in Table~\ref{table:tuning_sca}. 

\begin{table}[!t]
\centering
\caption{Comparison of \underline{Accuracy Scores} among our unoptimized \textsc{SCA0.1} and \textsc{Tuned SCA0.1} (kernel: 5$\rightarrow$ 7) in all 3 setups on CelebA and Medical MNIST datasets.}
\begin{tabular}{ p{1.3cm} p{3.5cm} p{3.0cm} p{3.0cm}}
\hline
 Dataset & Setup   & \textsc{SCA0.1} $\upuparrows$    &\textsc{Tuned SCA0.1} $\upuparrows$ \\
\hline
CelebA
   & \textsc{plug and play} & $0.726$	&$\mathbf{0.730}$	\\
& \textsc{end to end} & $0.748$	&$\mathbf{ 0.751}$	\\
& \textsc{split} & $0.745$	&$\mathbf{0.759}$\\ 

\hline
\hline

Medical& \textsc{plug and play} & $0.888$	&$\mathbf{0.899}$	\\
 MNIST& \textsc{end to end} & $ 0.888$	&$\mathbf{0.996}$\\
& \textsc{split} & $0.946$	&$\mathbf{0.967}$\\ 
\hline

\end{tabular}
\label{table:tuning_sca}
\end{table}

\label{ssec:computetime}
\begin{table}[h!]
\centering
 \caption{}
\begin{tabular}{cc}\\\toprule  
 $Model$  & \textsc{Time (sec)}   \\ \midrule
  \textsc{No-Defense} &10555.3	\\  \midrule
 \textsc{Gaussian-Noise} &12555.3\\ \midrule 
 \textsc{GAN} &15762.4\\  \midrule
  \mbox{Titcombe et al.\cite{titcombe2021practical}} &14390.2	\\  \midrule
\mbox{Gong et al.\cite{gong2023gan}}  &16061.8	\\  \midrule
\mbox{Gong et al.\cite{gong2023gan}}++  &17521.8	\\  \midrule
\mbox{Peng et al.\cite{peng2022bilateral}} &18921.2	\\ \midrule
\mbox{Hayes et al.\cite{hayes2023bounding}} &16923.9	\\ \midrule
\mbox{Wang et al.\cite{wang2021improving}} &15229.9	\\ \midrule

\textsc{Sparse-Standard} &12327.5\\ \midrule
\textsc{SCA0.1} & 17009.8\\ \midrule
\textsc{SCA0.25} & 17181.2\\ \midrule
\textsc{SCA0.5} & 17912.9\\ 

\bottomrule
 \end{tabular}
 \label{tab:computetimes}
 \end{table}

\section{Robustness of sparse coding layers: UMap}
\label{app:umap}
\begin{figure}[t]
\begin{subfigure}{.33\textwidth}
  \centering
  \includegraphics[width=3.0cm]{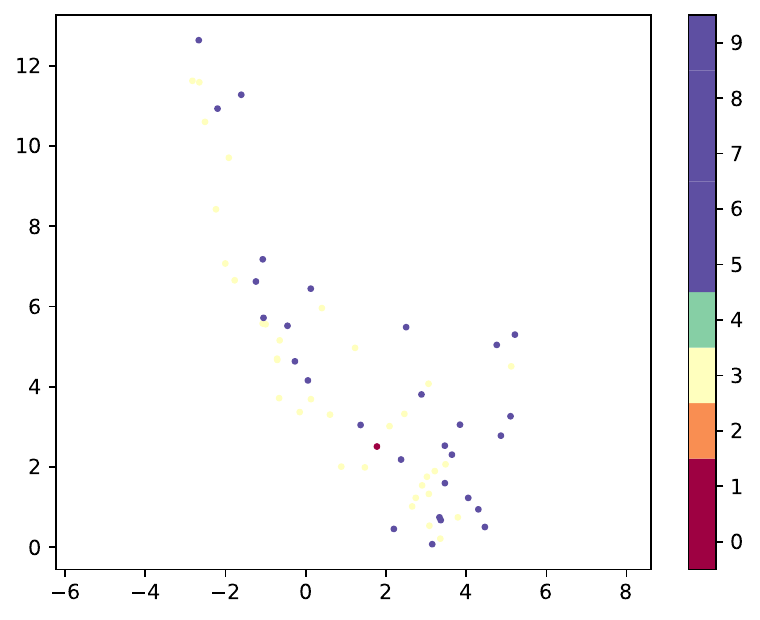}
  \caption{Linear Layer}
  \label{fig:dist_linear_one_cel}
\end{subfigure}%
\begin{subfigure}{.33\textwidth}
  \centering
  \includegraphics[width=3.0cm]{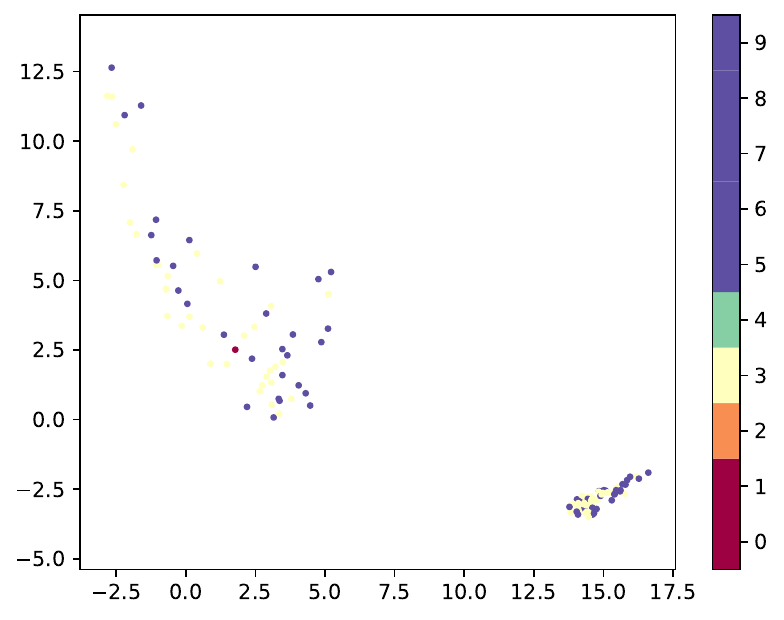}
  \caption{Convolution Layer}
  \label{fig:dist_cnn_one_cel}
\end{subfigure}
\begin{subfigure}{.33\textwidth}
  \centering
  \includegraphics[width=3.0cm]{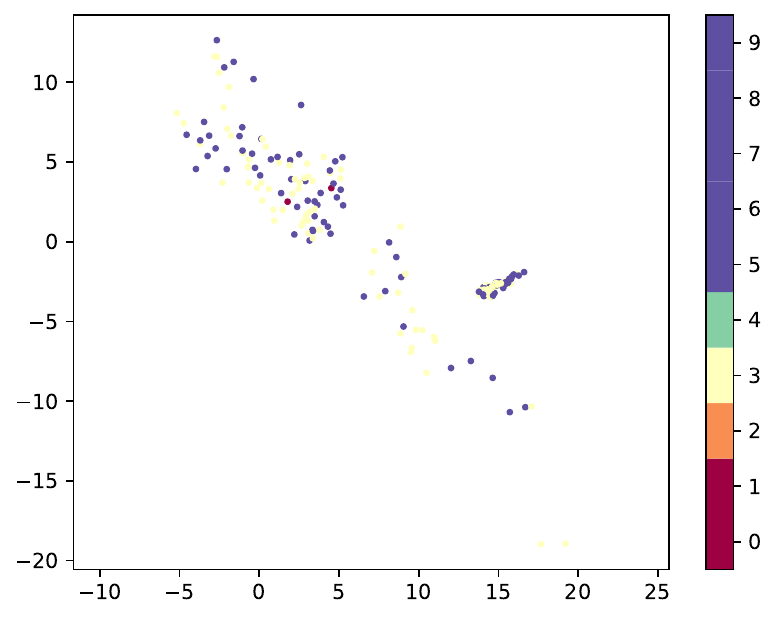}
  \caption{Sparse Coding Layer}
  \label{fig:dist_lca_one_cel}
\end{subfigure}
\begin{subfigure}{.33\textwidth}
  \centering
  \includegraphics[width=3.0cm]{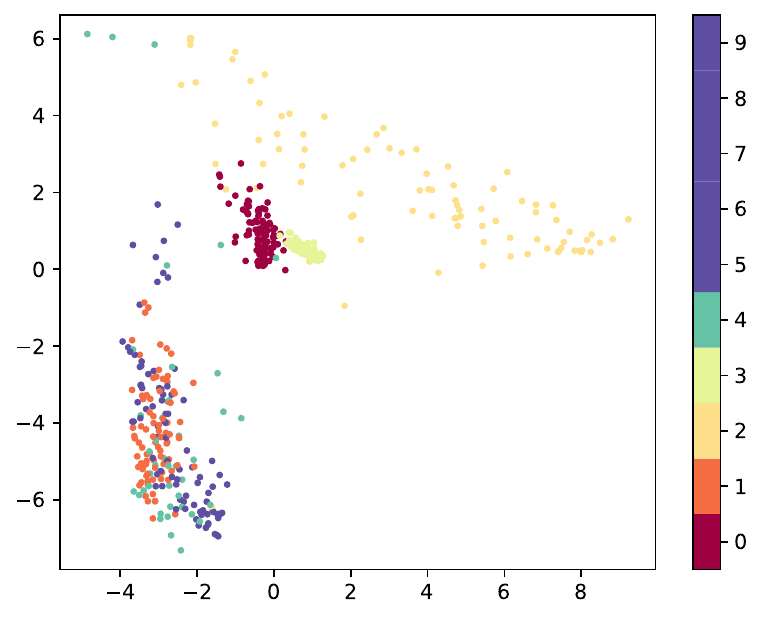}
  \caption{Linear Layer}
  \label{fig:dist_linear_mm}
\end{subfigure}%
\begin{subfigure}{.33\textwidth}
  \centering
  \includegraphics[width=3.0cm]{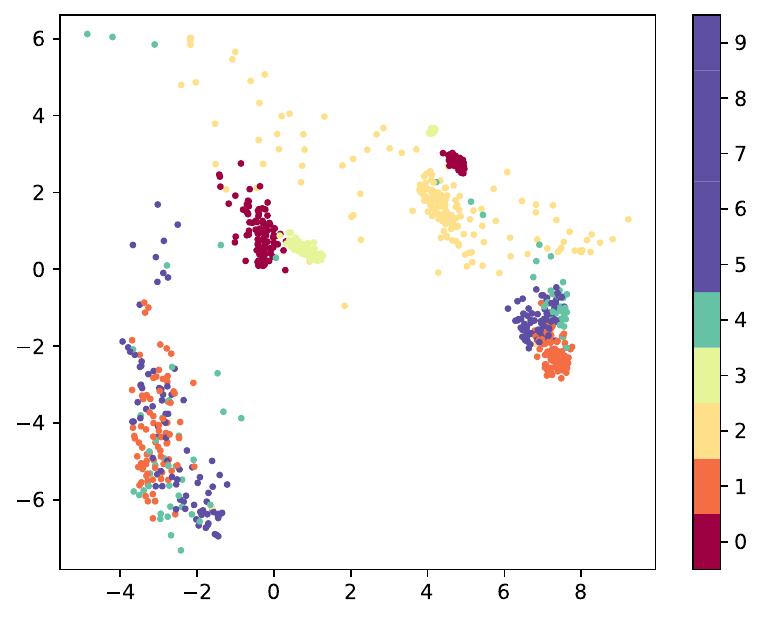}
  \caption{Convolution Layer}
  \label{fig:dist_cnn_mm}
\end{subfigure}
\begin{subfigure}{.33\textwidth}
  \centering
  \includegraphics[width=3.0cm]{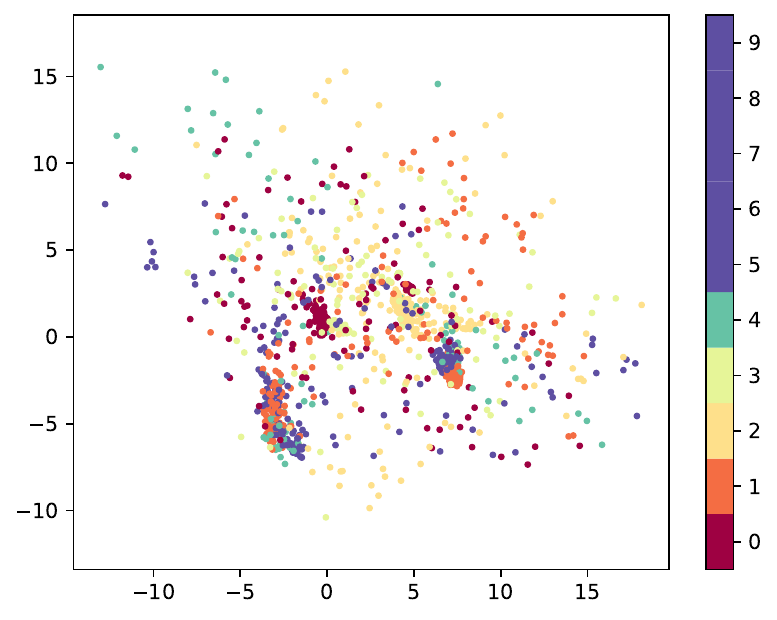}
  \caption{Sparse Coding Layer}
  \label{fig:dist_lca_mm}
\end{subfigure}
\caption{ UMap 2D projections of input images' features by class after $2$ linear layers, $2$ conv. layers, or $2$ sparse-coded layers on CelebA (top) \& Medical MNIST (bottom).}
\label{fig:dist_umap_mm_and_cel}
\end{figure}

In Figure~\ref{fig:dist_umap_mm_and_cel}, we present the UMap representation of linear, convolutional, and sparse coding layers on the other datasets, i.e., CelebA and Medical MNIST datasets. Observe that, the data points are more scattered in the sparse coding layer UMap (Figure~\ref{fig:dist_lca_one_cel} and Figure~\ref{fig:dist_lca_mm}) representations compared to the linear (Figure~\ref{fig:dist_linear_one_cel} and Figure~\ref{fig:dist_linear_mm}) and convolutional layers (Figure~\ref{fig:dist_cnn_one_cel} and Figure~\ref{fig:dist_cnn_mm}), which provide more robustness to models with sparse coding layers, i.e., our proposed \textsc{SCA}, against the privacy attacks.

\clearpage

\end{document}